\def\year{2020}\relax
\documentclass[letterpaper]{article} 
\usepackage{aaai20}  
\usepackage{times}  
\usepackage{helvet} 
\usepackage{courier}  
\usepackage[hyphens]{url}  
\usepackage{graphicx} 
\usepackage{graphicx}  
\nocopyright
\makeatletter
   \renewcommand\@biblabel[1]{}
\makeatother

\usepackage{amssymb}
\usepackage{amsmath}
\usepackage{booktabs}
\usepackage{enumerate}
\usepackage{algorithm}
\usepackage[noend]{algpseudocode}
\usepackage{subfig}
\usepackage{stackengine}
\newcommand{\norm}[1]{\left\lVert#1\right\rVert}
\DeclareMathOperator*{\argmax}{arg\,max}
\DeclareMathOperator*{\argmin}{arg\,min}
\makeatletter
    \newcommand*{\rom}[1]{\expandafter\@slowromancap\romannumeral #1@}
\makeatother
\newcommand{\lowerromannumeral}[1]{\romannumeral#1\relax}

\newlength{\tempdima}
\newcommand{\rowname}[1]
{\rotatebox{90}{\makebox[\tempdima][c]{#1}}}

\title{Capsule Routing via Variational Bayes}
\author{\Large \textbf{Fabio De Sousa Ribeiro, Georgios Leontidis, Stefanos Kollias} \\ 
Machine Learning Group \\ 
School of Computer Science, University of Lincoln, UK\\
\texttt{\{fdesousaribeiro,gleontidis,skollias\}@lincoln.ac.uk}}

\begin{document}
\maketitle
\begin{abstract}
Capsule networks are a recently proposed type of neural network shown to outperform alternatives in challenging shape recognition tasks. In capsule networks, scalar neurons are replaced with capsule vectors or matrices, whose entries represent different properties of objects. The relationships between objects and their parts are learned via trainable viewpoint-invariant transformation matrices, and the presence of a given object is decided by the level of agreement among votes from its parts. This interaction occurs between capsule layers and is a process called \textit{routing-by-agreement}. In this paper, we propose a new capsule routing algorithm derived from Variational Bayes for fitting a mixture of transforming gaussians, and show it is possible transform our capsule network into a Capsule-VAE. Our Bayesian approach addresses some of the inherent weaknesses of MLE based models such as the \textit{variance-collapse} by modelling uncertainty over capsule pose parameters. We outperform the state-of-the-art on smallNORB using $\simeq$50\% fewer capsules than previously reported, achieve competitive performances on CIFAR-10, Fashion-MNIST, SVHN, and demonstrate significant improvement in MNIST to affNIST generalisation over previous works.\footnote{https://github.com/fabio-deep/Variational-Capsule-Routing}
\end{abstract}
\begin{figure}[t]
    \centering
    \includegraphics[width=\columnwidth]{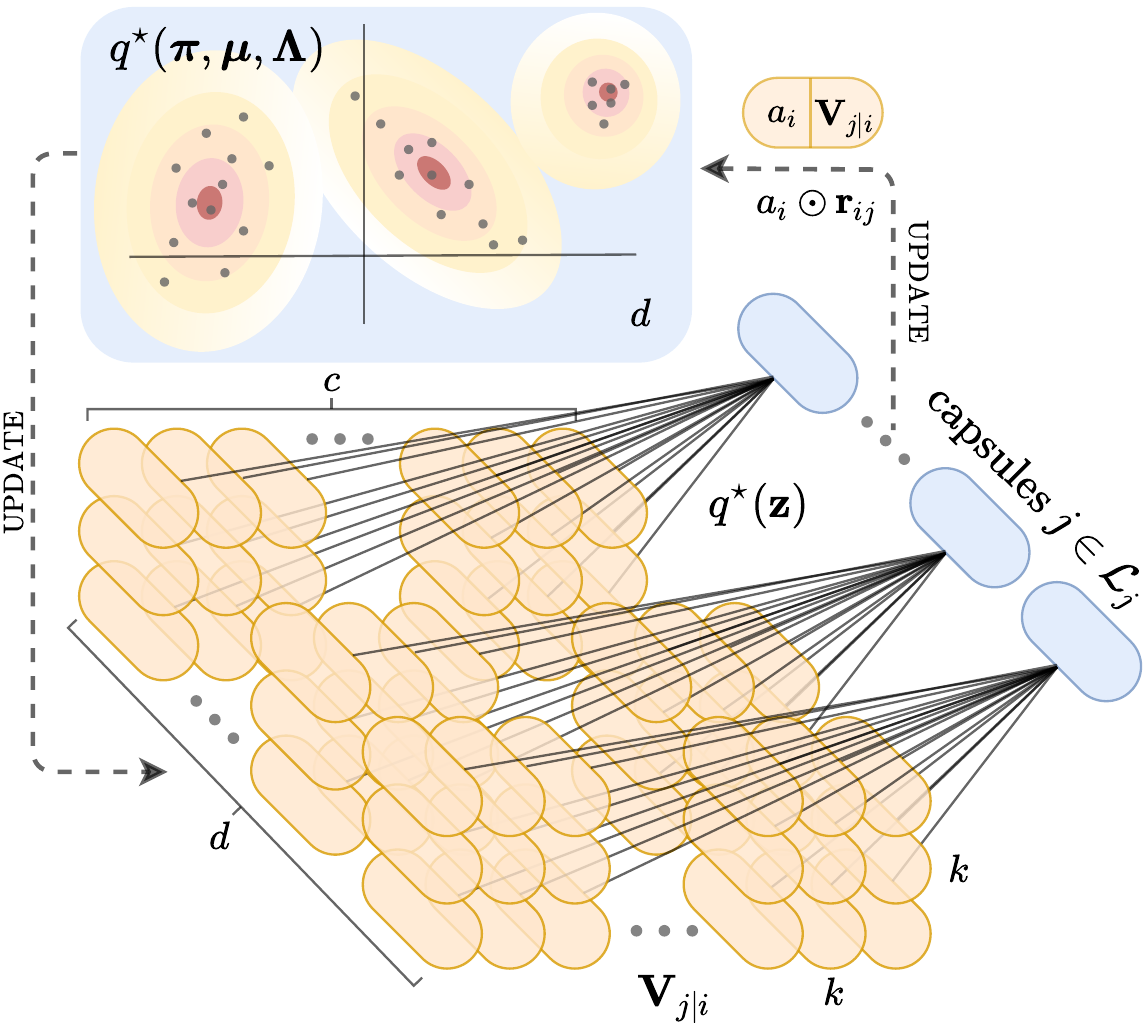}
 \caption{Depiction of Variational Bayes (VB) routing between adjacent capsule layers: with lower layer capsules $i\in\mathcal{L}_i$ (orange) and higher layer capsules $j\in\mathcal{L}_j$ (blue).}
\label{plots}
\end{figure}
\section{Introduction}
\begin{figure*}[t]
    \centering
    \includegraphics[width=2.11\columnwidth]{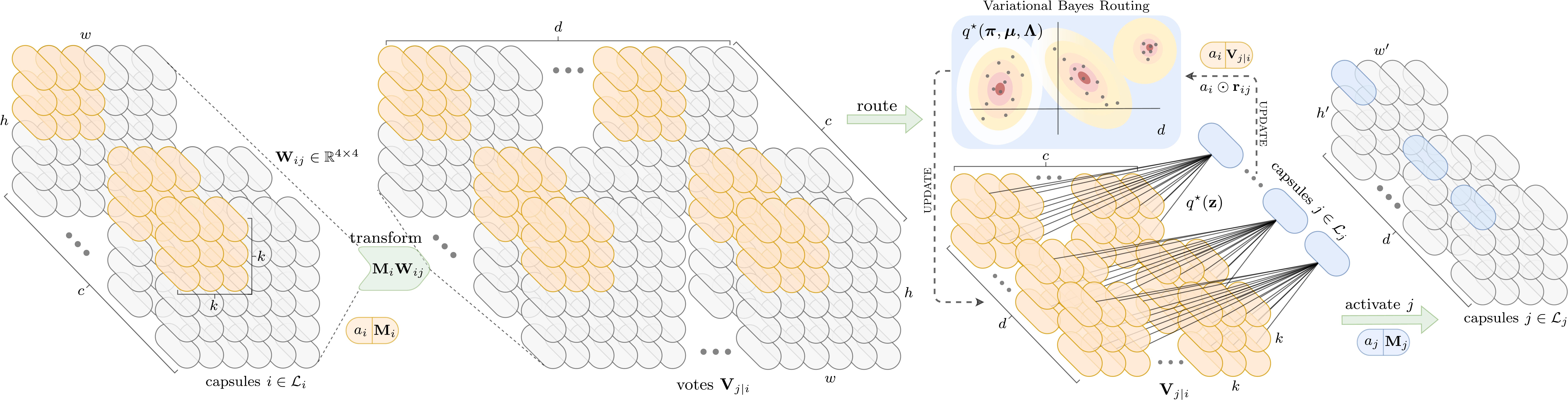}
 \caption{Architectural depiction of our capsule network with Variational Bayes routing between convolutional capsule layers. Each capsule has an activation probability $a$ and a pose matrix $\mathbf{M}\in\mathbb{R}^{4\times4}$. Parent capsules $j$ (blue) only receive votes from child capsules $i$ (orange) within their receptive field. $c$ and $d$ denote the number of child and parent capsule types respectively.}
\label{big_diagram}
\end{figure*}

Capsule networks are a recently proposed method of learning part-whole relationships between observed entities in data, by using groups of neurons known as capsules. These entities could be anything that possesses a consistent underlying structure across viewpoints. Capsules attempt to encode intrinsic viewpoint-invariant properties, and learn to adjust instantiation parameters as the entity varies across its appearance manifold~\cite{hinton2011transforming}. CapsNets have shown to outperform standard Convolutional Neural Networks (CNNs) in specific tasks involving shape recognition and overlapping digit segmentation. These tasks are difficult for standard CNNs, as they struggle to exploit the frame of reference humans impose on objects, and thus often fail to generalise knowledge to novel viewpoints. Although this drawback can often be mitigated by data augmentation during training, it does not address the underlying issue directly. Nonetheless, CNNs perform remarkably well in practice, partly because they make structural assumptions that ring true with natural images. Capsules extend this rationale by assuming objects are composed of object parts, and if we learn part-whole relationships perfectly then we can better generalise to novel viewpoints and affine transformations. In CNNs, the convolution operator and sparse weight sharing provides the useful property of equivariance under translation, enabling efficient spatial transfer of knowledge. CapsNets retain these benefits and only do away with pooling operations in favour of learning more robust representations for disentangling factors of variation with \textit{routing-by-agreement}. Although promising, CapsNets remain underexplored, and few works thus far have proposed algorithmic improvements to the original formulations. In this paper, we propose a new capsule routing algorithm for fitting a mixture of transforming gaussians via Variational Bayes, which offers increased training stability, flexibility and performance. 
\paragraph{Capsule Networks}\label{sec: Capsule Networks}
CapsNets are composed of at least one layer of capsules in which capsules $i$ from a lower layer $\mathcal{L}_{i}$ (children) are routed to capsules $j$ in a higher layer $\mathcal{L}_{j}$ (parents). Each layer contains multiple lower capsules, each of which has a pose matrix $\mathbf{M}_{i} \in \mathbb{R}^{4\times4}$ of instantiation parameters and activation probability $a_i$ (see Figure~\ref{big_diagram}). The pose matrix may learn to encode the relationship of an entity to the viewer, and the activation probability $a_{i}$ represents its presence. Each lower level capsule uses its pose matrix $\mathbf{M}_{i}$ to posit a vote for what the pose of a higher level capsule should be, by multiplying it with a trainable viewpoint-invariant transformation weight matrix 
\begin{equation}
\mathbf{V}_{j|i} = \mathbf{M}_{i}\mathbf{W}_{ij},
\end{equation}
where $\mathbf{V}_{j|i}$ denotes the vote coming from capsules $i$ to capsule $j$, and $\mathbf{W}_{ij} \in \mathbb{R}^{4\times4}$ is the trainable transformation matrix. To compute the pose matrix $\mathbf{M}_{j}$ of any higher level capsule $j$ we can simply take a weighted mean of the votes it received from capsules in $\mathcal{L}_{i}$ as in EM routing~\cite{hinton2018matrix}: $\mathbf{M}_{j} = 1/R_j \sum_i{\mathbf{V}_{j|i}}R_{ij}$, where $R_{ij}$ represents the posterior responsibilities of each capsule $j$ for capsules $i$, and $R_j = \sum_i{R_{ij}}$. These routing coefficients can be tuned via a variant of the EM algorithm for Gaussian Mixtures, and are updated according to the agreement between $\mathbf{V}_{j|i}$ and $\mathbf{M}_{j}$, which in Dynamic routing~\cite{sabour2017dynamic} for example, is simply the scalar product between capsule vectors and can be trivially extended to matrices with $\norm{\mathbf{M}_{j} - \mathbf{V}_{j|i}}_{F}$. Lastly, a parent capsule $j$ is only activated if there is a measurably high agreement among the votes $\mathbf{V}_{j|i}$ from child capsules $i$ for its pose matrix $\mathbf{M}_j$, which forms a tight cluster in $\mathbb{R}^D$.
\paragraph{Motivation \& Contributions}
In this paper, we propose a new capsule routing algorithm derived from Variational Bayes. We show that our probabilistic approach provides advantages over previous routing algorithms, including more flexible control over capsule complexity by tuning priors to induce sparsity, and reducing the well known \textit{variance-collapse} singularities inherent to MLE based mixture models such as EM. Contextually, these singularities occur in part due to the single parent assumption--whereby a parent capsule (gaussian cluster) can claim sole custody of a child capsule (datapoint), yielding infinite likelihood and zero variance. This leads to overfitting and unstable training. By modelling uncertainty over the capsule parameters as well as the routing weights, we can avoid these singularities in a principled way, without adding arbitrary constants of minimum variance to ensure numerical stability, which can affect performance in EM. Furthermore, we provide some insight into capsule network training for practitioners including weight initialisation and normalisation schemes that improve training performance. Lastly, we show it's possible to transform our capsule network into a Capsule-VAE by sampling latent code from capsule parameter approximate posteriors. We outperform the state-of-the-art on smallNORB using $\simeq$50\% fewer capsules than previously reported, achieve highly competitive performances on CIFAR-10, Fashion-MNIST, SVHN, and demonstrate significant improvement in MNIST to affNIST generalisation over previous works.
\section{Variational Bayes Capsule Routing}\label{sec: Variational Bayes Routing}
Next, we briefly outline some necessary background on Variational Inference (VI), before contextualising some of these ideas with our proposed capsule routing algorithm.
\subsection{Variational Inference}\label{subsec: Variational Inference}
\paragraph{The Evidence Lower Bound} Let $\mathbf{x}$ denote the observed data, $\mathbf{z}$ denote latent variables associated with $\mathbf{x}$, and let $\boldsymbol{\theta}$ represent some model parameters. Typically we'd like to infer the unknown latent variables, by evaluating the conditional $p(\mathbf{z}|\mathbf{x},\boldsymbol{\theta})$ which is the posterior on $\mathbf{z}$. However, this distribution cannot be computed for most complex models due to the intractability of a normalising integral. VI provides an elegant solution to posterior inference by posing it as an optimisation problem. We approximate the posterior $p(\mathbf{z}|\mathbf{x},\boldsymbol{\theta})$ by choosing a variational distribution over the latent variables $q_{\phi}(\mathbf{z})$ from a tractable family, with its own variational parameters $\phi$. We can measure the quality of our approximation via the Kullback-Leibler (KL) divergence $\mathrm{KL}[q_{\phi}(\mathbf{z}) \ || \ p(\mathbf{z}|\mathbf{x},\boldsymbol{\theta})]$ between the two distributions, which can be minimised via the variational parameters $\phi$
\begin{equation}
    \widehat{\phi} = \argmin_{\phi} \ \mathbb{E}_{q_{\phi}(\mathbf{z})} [\log q_{\phi}(\mathbf{z}) - \log p(\mathbf{z}|\mathbf{x},\boldsymbol{\theta})].
\end{equation}
However, since $p(\mathbf{z}|\mathbf{x},\boldsymbol{\theta})$ is unknown we cannot minimise the KL directly, so instead we maximise the variational lower bound (ELBO) on the log marginal likelihood 
\begin{equation}
    \log p(\mathbf{x}|\boldsymbol{\theta}) = \text{KL}[q_{\phi}(\mathbf{z}) \ || \ p(\mathbf{z}|\mathbf{x},\boldsymbol{\theta})] \ + \ \mathcal{L}_{\text{ELBO}}(q_{\phi}(\mathbf{z})),
\end{equation}
where the ELBO can be derived using Jensen's inequality $\log(\mathbb{E}[X]) \geq \mathbb{E}[\log(X)]$ applied to $\log p(\mathbf{x}|\boldsymbol{\theta})$ giving 
\begin{equation}
\begin{aligned}
    \log p(&\mathbf{x}|\boldsymbol{\theta})\geq \mathcal{L}_{\text{ELBO}}(q_{\phi}(\mathbf{z})) = \\ &= \mathbb{E}_{q_{\phi}(\mathbf{z})} [\log p(\mathbf{x},\mathbf{z}|\boldsymbol{\theta})] - \mathbb{E}_{q_{\phi}(\mathbf{z})} [\log q_{\phi}(\mathbf{z})].  
\end{aligned}
\end{equation}
Here we use the joint $\log p(\mathbf{x},\mathbf{z}|\boldsymbol{\theta})$ which is tractable, rather than the unknown posterior $\log p(\mathbf{z}|\mathbf{x},\boldsymbol{\theta})$. Recall that from the product rule of probability we simply have that $p(\mathbf{x},\mathbf{z}|\boldsymbol{\theta}) = p(\mathbf{z}|\mathbf{x},\boldsymbol{\theta})p(\mathbf{x}|\boldsymbol{\theta})$. Given that the log marginal likelihood of the data $\log p(\mathbf{x}|\boldsymbol{\theta})$ is always negative and is independent of $q_{\phi}(\mathbf{z})$, maximising the ELBO is therefore equivalent to minimising the KL divergence.
\paragraph{Mean Field}
A popular way of performing VI is to posit a factorised form of the approximating family of distributions $q_\phi(\mathbf{z})$, such that each variable is assumed to be independent
\begin{equation}
    p(\mathbf{z}|\mathbf{x},\boldsymbol{\theta}) \approx q_\phi(\mathbf{z}) = \prod_{i=1}^{N} q_{\phi_i}(\mathbf{z}_{i}), \quad \sum_{\mathbf{z}_{i}}q_{\phi_i}(\mathbf{z}_i) = 1.    
\end{equation}
Recall that the log marginal is given by $\log p(\mathbf{x}|\boldsymbol{\theta}) = \log \sum_\mathbf{z}p(\mathbf{x},\mathbf{z}|\boldsymbol{\theta})$, and therefore the factorised objective to be maximised can be written in the following form
\begin{equation}
\begin{aligned}
    \argmax_{q_{\phi_i}(\mathbf{z}_i) \in q_\phi(\mathbf{z})} \sum_{i=1}^{N} 
     \mathbb{E}_{q_{\phi_i}(\mathbf{z}_i)} [\log& p(\mathbf{x}_i,\mathbf{z}_i|\boldsymbol{\theta})] \\ &- \mathbb{E}_{q_{\phi_i}(\mathbf{z}_i)} [\log q_{\phi_i}(\mathbf{z}_i)].
\end{aligned}
\end{equation}
\subsection{Variational Bayes for a Mixture of Transforming Gaussians}\label{subsec: Variational Bayes for a Mixture of Transforming Gaussians}
\paragraph{Relation to Clustering} Capsule routing naturally resembles clustering logic. This is reflected in the fact that any higher layer parent capsule $j$ (cluster) is composed of, and receives votes from, many lower layer child capsules $i$ (data points) within its receptive field (see Figure~\ref{big_diagram} for intuition). 

However, capsule routing does differ from regular clustering substantially, as every cluster has its own learnable viewpoint-invariant transformation matrix $\mathbf{W}_{ij}$ with which it transforms its data points, and predictions are made by measuring similarity among them. Therefore, each cluster sees a different view of the data, and the algorithm converges much faster since it's easier to break symmetry compared to simply initialising the gaussian clusters with different means~\cite{hinton2018matrix}. Next we propose our capsule routing algorithm borrowing some ideas from~\cite{bishop2006pattern}, and begin by picking up from our general description of capsule networks in section~\ref{sec: Capsule Networks}. 
\paragraph{Proposed Method}
Let $\mathbf{v}_{j|i} \in \mathbb{R}^{D}$ denote a vectorised version of the $4\text{x}4$ votes $\mathbf{V}_{j|i}$ matrix, and let $\boldsymbol{\mu}_j \in \mathbb{R}^{D}$ denote a vectorised version of capsule $j$'s $4\text{x}4$ pose matrix $\mathbf{M}_j$, where $D=16$. Assuming independence, consider the log likelihood function maximised in a Gaussian Mixture Model (GMM), applied to routing capsules $i$ from a lower layer to capsules $j$ in a higher layer
\begin{equation}
    \log p(\mathbf{v}|\boldsymbol{\pi},\boldsymbol{\mu},\boldsymbol{\Lambda}) = \sum_{i \in \mathcal{L}_i} \log \sum_{j\in \mathcal{L}_j}\pi_j \mathcal{N}(\mathbf{v}_{j|i}|\boldsymbol{\mu}_j,\boldsymbol{\Lambda}_j^{-1}).
\end{equation}
In EM routing, point estimates of the parameters $\boldsymbol{\mu}_j$ and $\mathrm{diag}(\boldsymbol{\Lambda}_{j})$ are computed in the M-step, and the routing probabilities $R_{ij}$ are evaluated in the E-step. The mixing coefficients $\pi_j$ however, are replaced with activations $a_j$ which represent the probability of cluster $j$ being switched on, and are computed by a shifting logistic non-linearity. The $a_j$'s play the role of the mixing proportions but $\sum_j \mathbf{a}_j \neq 1$. Recall from section~\ref{sec: Capsule Networks} that the votes play the roles of the data points and are computed as $\mathbf{V}_{j|i} = \mathbf{M}_{i}\mathbf{W}_{ij}$, using different transformation matrices 
$\mathbf{W}_{ij}$ for each capsule $j$.

In order to model uncertainty over the capsule parameters in our algorithm, we place conjugate priors over $\boldsymbol{\pi}$, $\boldsymbol{\mu}$ and $\boldsymbol{\Lambda}$. Our model's generative process for any lower layer capsule $i$'s vectorised pose $\boldsymbol{\mu}_{i:\mathcal{L}_i}$ can be derived from the following
\begin{equation}
\begin{aligned}
    \mathbf{v}_{j|i} \ |\ \mathbf{z}_i = j
    &\sim\mathcal{N}(\boldsymbol{\mu}_j,\boldsymbol{\Lambda}_j^{-1}) \\
    \mathbf{z}_i &\sim\mathrm{Cat}(\mathbf{z}_i|\boldsymbol{\pi}) \\
    \boldsymbol{\pi} \ | \ \boldsymbol{\alpha}_0 \ &\sim\mathrm{Dir}(\boldsymbol{\alpha}_0) \\
    \boldsymbol{\mu}_j \ | \ \mathbf{m}_0, \kappa_0, \boldsymbol{\Lambda}_j &\sim\mathcal{N}(\mathbf{m}_0, (\kappa_0\boldsymbol{\Lambda}_j)^{-1}) \\
    \boldsymbol{\Lambda}_j \ | \ \boldsymbol{\Psi}_0, \nu_0 &\sim\mathrm{Wi}(\boldsymbol{\Psi}_0, \nu_0), \\
\end{aligned}
\end{equation}
and $\boldsymbol{\mu}_{i}$ can be retrieved by simply inverting the vectorised vote transformation $\boldsymbol{\mu}_i = \mathbf{w}_{ij}^{-1}\mathbf{v}_{j|i}$. The joint distribution of the model factorises as $p(\mathbf{v},\mathbf{z}, \boldsymbol{\pi}, \boldsymbol{\mu}, \boldsymbol{\Lambda}) =  p(\mathbf{v}|\mathbf{z},\boldsymbol{\mu}, \boldsymbol{\Lambda}) p(\mathbf{z}|\boldsymbol{\pi}) p(\boldsymbol{\pi}) p(\boldsymbol{\mu}|\boldsymbol{\Lambda}) p(\boldsymbol{\Lambda})$, where the latent variables $\mathbf{z}$ are a collection of $\mathcal{L}_i$ one-hot vectors denoting the cluster assignments of each of the lower capsules votes $\mathbf{v}_{j|i}$, to their corresponding higher capsules' gaussians. 
Following from the VI discussion in section~\ref{subsec: Variational Inference}, we approximate the posterior $p(\mathbf{z},\boldsymbol{\pi},\boldsymbol{\mu},\boldsymbol{\Lambda}|\mathbf{v}) \propto p(\mathbf{v}|\mathbf{z},\boldsymbol{\mu},\boldsymbol{\Lambda}) p(\mathbf{z}|\boldsymbol{\pi})p(\boldsymbol{\pi})p(\boldsymbol{\mu},\boldsymbol{\Lambda})$ with a factorised variational distribution
\begin{equation}
    p(\mathbf{z},\boldsymbol{\pi},\boldsymbol{\mu},\boldsymbol{\Lambda}|\mathbf{v}) \approx q(\mathbf{z})q(\boldsymbol{\pi}) \prod_{j \in \mathcal{L}_j} q(\boldsymbol{\mu}_j, \boldsymbol{\Lambda}_j),
\end{equation}
and we choose conjugate priors that factor in the following standard form as in Bayesian Gaussian Mixtures
\begin{equation}
\begin{aligned}
    p(&\boldsymbol{\pi})p(\boldsymbol{\mu},\boldsymbol{\Lambda}) = 
    \mathrm{Dir}(\boldsymbol{\pi}|\boldsymbol{\alpha}_{0}) \\ &\times \prod_{j\in \mathcal{L}_j} {\mathcal{{N}} \big(\boldsymbol{\mu}_j|\mathbf{m}_0, (\kappa_0\boldsymbol{\Lambda}_j)^{-1}\big)} \mathrm{Wi}(\boldsymbol{\Lambda}_j|\boldsymbol{\Psi}_{0},\nu_{0}).
\end{aligned}    
\end{equation}
To parameterise diagonal precisions in practice, we simply let $\boldsymbol{\lambda}_j \in \mathbb{R}^{D}$ represent the diagonal entries of $\boldsymbol{\Lambda}_j$, and replace the Gaussian-Wishart prior with Gaussian-Gamma priors over each diagonal entry $\lambda_{j}^d$ as follows
\begin{equation}\label{eq: Gaussian-Gamma}
\begin{aligned}
    p(&\boldsymbol{\mu}|\boldsymbol{\lambda})p(\boldsymbol{\lambda})= \\ & \prod_{j\in \mathcal{L}_j} \prod_{d=1}^{D} \mathcal{N}(\mu_{j}^d|m_{0}, (\kappa_{0}\lambda_{j}^d)^{-1})\mathrm{Ga}(\lambda_{j}^d|s_{0},\nu_{0}).
\end{aligned}
\end{equation}
\begin{algorithm*}[t]
\footnotesize
\caption{Variational Bayes Capsule Routing}\label{algo}
\begin{algorithmic}[1]
\Function{VB Routing}{$\mathbf{a}_i,\mathbf{v}_{j|i}$}\Comment{Input votes and activations from preceding capsule layer $\mathcal{L}_i$}
\State {Initialise routing weights}{$\ \forall \ i$,$j$ : $\mathbf{r}_{ij} \gets 1/\mathcal{L}_j$}
\State {Initialise capsule priors}{$\ \forall \ j$ : $\alpha_0, \mathbf{m}_0, \kappa_0, \mathbf{S}_0, \nu_0$}
\For{$n$ \text{iterations}}
\State \Call{Update Suff. Stats}{}
\State \Call{Update $q^{\star}(\boldsymbol{\pi},\boldsymbol{\mu},\boldsymbol{\Lambda})$}{}
\State \Call{Update $q^{\star}(\mathbf{z})$}{}
\EndFor
\State{$\mathbf{a}_j \gets \mathrm{sigmoid\big(\beta_a - (\beta_u + \mathbb{E}[\ln\pi_j] + \mathbb{E}[\ln\det(\boldsymbol{\Lambda}_j)]) \odot \mathbf{r}_j}\big)$}\Comment{Activate using approximate $\mathbb{H}[q^{\star}(\boldsymbol{\mu}_j, \boldsymbol{\Lambda}_j)]$}
\State \textbf{return} $\mathbf{a}_j, \mathbf{m}_j$
\EndFunction
\Function{Update Suff. Stats}{$\mathbf{a}_i,\mathbf{v}_{j|i},\mathbf{r}_{ij}$}\Comment{Calculate sufficient statistics of incoming votes $\mathbf{v}_{j|i}$}
\State {$\mathbf{r}_{ij} \gets \mathbf{r}_{ij} \odot \mathbf{a}_i$}
\State {$\mathbf{r}_{j} \gets \sum_i \mathbf{r}_{ij}$}
\State {$\widetilde{\mathbf{v}}_{j} \gets 1/\mathbf{r}_j \sum_i \mathbf{r}_{ij} \mathbf{v}_{j|i}$}
\State {$\mathbf{S}_{j} \gets \sum_i \mathbf{r}_{ij} (\mathbf{v}_{j|i} - \widetilde{\mathbf{v}}_{j})(\mathbf{v}_{j|i} - \widetilde{\mathbf{v}}_{j})^{T}$}
\EndFunction
\Function{Update $q^{\star}(\boldsymbol{\pi},\boldsymbol{\mu},\boldsymbol{\Lambda})$}{}\Comment{Update capsule pose parameter distributions}
\State {$\alpha_{j} \gets \alpha_{0} + \mathbf{r}_j \ , \ \kappa_{j} \gets \kappa_{0} + \mathbf{r}_j \ , \  \nu_{j} \gets \nu_{0} + \mathbf{r}_j$}
\State {$\mathbf{m}_{j} \gets (\mathbf{r}_{j} \widetilde{\mathbf{v}}_{j} + \kappa_{0}\mathbf{m}_0)\kappa_{j}^{-1}$}
\State {$\boldsymbol{\Psi}_{j}^{-1} \gets \boldsymbol{\Psi}_{0}^{-1}  + \mathbf{S}_{j} +
\kappa_0\mathbf{r}_j\kappa_j^{-1} (\widetilde{\mathbf{v}}_{j} - \mathbf{m}_0)(\widetilde{\mathbf{v}}_{j} - \mathbf{m}_0)^{T}$}
\State {$\ln\det(\boldsymbol{\Psi}_{j}) \gets -2\mathrm{trace}(\ln\mathrm{Cholesky}(\boldsymbol{\Psi}_{j}^{-1}))$}
\EndFunction
\Function{Update $q^{\star}(\mathbf{z})$}{}\Comment{Update posterior routing responsibilities}
\State {$\mathbb{E}[\ln\pi_j] \gets \psi(\alpha_j) - \psi(\sum_j \alpha_j)$}
\State {$\mathbb{E}[\ln\det(\boldsymbol{\Lambda}_j)] \gets D\ln2 + \ln\det(\boldsymbol{\Psi}_{j}) + \sum_{i=0}^{D-1}\psi\big((\nu_j - i) /2 \big)$}
\State {$\mathbb{E}[\mathcal{D}_{\mathrm{maha}}(\mathbf{v}_{j|i}, \boldsymbol{\mu}_j)] \gets D\kappa_j^{-1} + \nu_j(\mathbf{v}_{j|i} - \mathbf{m}_j)^{T}\boldsymbol{\Psi}_j(\mathbf{v}_{j|i} - \mathbf{m}_j)$}
\State {$\ln \mathbf{p}_{j} \gets \mathbb{E}[\ln\det(\boldsymbol{\Lambda}_j)]/2 -\mathbb{E}[\mathcal{D}_{\mathrm{maha}}(\mathbf{v}_{j|i}, \boldsymbol{\mu}_j)]/2$}
\State {$\mathbf{r}_{ij} \gets \mathrm{softmax}\big(\mathbb{E}[\ln\pi_j] + \ln \mathbf{p}_{j}\big)$}\Comment{Normalise over capsules $j \in \mathcal{L}_j$}
\EndFunction
\end{algorithmic}
\end{algorithm*}
In order to perform routing, we simply iterate between optimising parent capsule parameter distributions $q^{\star}(\boldsymbol{\pi},\boldsymbol{\mu},\boldsymbol{\Lambda})$ using the responsibilities over child capsules fixed, and evaluating the new expected responsibilities $q^{\star}(\mathbf{z})$ using the current distributions over parent capsule parameters fixed. See Algorithm~\ref{algo} for the standard closed-form update equations, which assume the same functional form as the priors through conjugacy, and for further details refer to~\cite{bishop2006pattern}.
\begin{table*}[t]
\caption{Test error rate comparisons with CapsNet literature. ($\cdot$) denotes ensemble size, and ($\dagger$) denotes our EM implementation.}
\footnotesize
    \begin{tabular}{lcrcrcrcr}
        \toprule
        & \multicolumn{2}{c}{\textbf{smallNORB}} & \multicolumn{2}{c}{\textbf{Fashion-MNIST}} & \multicolumn{2}{c}{\textbf{SVHN}} & \multicolumn{2}{c}{\textbf{CIFAR-10}} \\
        \textbf{Method} & Error ($\%$) & Param & Error ($\%$) & Param & Error ($\%$) & Param & Error ($\%$) & Param \\
        \midrule
        HitNet~\scriptsize{(Deliège et al. \citeyear{deliege2019effective})} & \multicolumn{1}{c}{-} & \multicolumn{1}{c}{-} & 7.7$\%$  & $\simeq$8.2M & 5.5$\%$ & $\simeq$8.2M & 26.7$\%$ & $\simeq$8.2M \\
        DCNet~\scriptsize{\cite{phaye2018dense}} & 5.57$\%$ & 11.8M & 5.36$\%$ & 11.8M & 4.42$\%$ & 11.8M & 17.37$\%$ & 11.8M \\
        MS-Caps~\scriptsize{\cite{xiang2018ms}} & \multicolumn{1}{c}{-} & \multicolumn{1}{c}{-} & 7.3$\%$ & 10.8M & \multicolumn{1}{c}{-} & \multicolumn{1}{c}{-} & 24.3$\%$ & 11.2M \\
        Dynamic~\scriptsize{\cite{sabour2017dynamic}} & 2.7$\%$ & 8.2M & \multicolumn{1}{c}{-} & \multicolumn{1}{c}{-} & 4.3$\%$ & $\simeq$1.8M & 10.6$\%$ & 8.2M \scriptsize{(7)} \\
        Nair \textit{et al}.~\scriptsize{\cite{nair2018pushing}} & \multicolumn{1}{c}{-} & \multicolumn{1}{c}{-} & 10.2$\%$ & 8.2M & 8.94$\%$ & 8.2M & 32.47$\%$ & 8.2M \\
        FRMS~\scriptsize{\cite{zhang2018fast}} & 2.6$\%$ & 1.2M & 6.0$\%$ & 1.2M & \multicolumn{1}{c}{-} & \multicolumn{1}{c}{-} & 15.6$\%$ & 1.2M \\
        MaxMin~\scriptsize{\cite{zhao2019capsule}} & \multicolumn{1}{c}{-} & \multicolumn{1}{c}{-} & 7.93$\%$ & $\simeq$8.2M & \multicolumn{1}{c}{-} & \multicolumn{1}{c}{-} & 24.08$\%$ & $\simeq$8.2M \\
        KernelCaps~\scriptsize{\cite{killian2019kernelized}} & \multicolumn{1}{c}{-} & \multicolumn{1}{c}{-} & \multicolumn{1}{c}{-} & \multicolumn{1}{c}{-} & 8.6$\%$ & $\simeq$8.2M & 22.3$\%$ & $\simeq$8.2M \\
        FREM~\scriptsize{\cite{zhang2018fast}} & 2.2$\%$ & 1.2M & 6.2$\%$ & 1.2M & \multicolumn{1}{c}{-} & \multicolumn{1}{c}{-} & 14.3$\%$ & 1.2M \\
        EM-Routing~\scriptsize{\cite{hinton2018matrix}} & 1.8$\%$ & 310K & \multicolumn{1}{c}{-} & \multicolumn{1}{c}{-} & \multicolumn{1}{c}{-} & \multicolumn{1}{c}{-} & 11.9$\%$ & $\simeq$460K \\
        \midrule
        VB-Routing: $\{64, 8, 
16, 16, d_4\}$ & 1.93$\%$ & 142K & 5.46$\%$ & 145K & 4.75$\%$ & 145K & 13.1$\%$ & 145K  \\
        VB-Routing: $\{64, 16, 
32, 32, d_4\}$ & 1.84$\%$ & 318K & 5.61$\%$ & 323K & \textbf{3.9}$\%$\scriptsize{$\pm.06$} & \textbf{323K} & \textbf{11.2}$\%$\scriptsize{$\pm.09$} & \textbf{323K}  \\
        \ \ vs. EM-Routing$^{\dagger}$: $\{64, 16, 
32, 32, d_4\}$ & - & \multicolumn{1}{c}{-} &  - & \multicolumn{1}{c}{-} & 5.17\% & 323K & 12.26\% & 323K  \\
        VB-Routing: $\{64, 16, 
16, 16, d_4\}$ & \textbf{1.6}$\%$\scriptsize{$\pm.06$} & \textbf{169K} &  \textbf{5.2}$\%$\scriptsize{$\pm.07$} & \textbf{172K} & 4.18\% & 172K & 12.4$\%$ & 172K  \\
        \ \ vs. EM-Routing$^{\dagger}$: $\{64, 16, 
16, 16, d_4\}$ & 1.97$\%$ & 169K &  6.14\% & 172K & - & \multicolumn{1}{c}{-} & - & \multicolumn{1}{c}{-} \\
        \bottomrule
    \end{tabular}
    \label{results table}
\end{table*}
\paragraph{Agreement \& Activation}
We propose to measure agreement between the votes from lower capsules $i$ using the differential entropy of a higher capsule $j$'s Gaussian-Wishart variational posterior distribution $q^{\star}(\boldsymbol{\mu}_j, \boldsymbol{\Lambda}_j)$. Firstly, the differential entropy of a multivariate gaussian distributed random variable $\mathbf{x}$ is by definition given by
\begin{equation}
\begin{aligned}
    \mathbb{H}[\mathbf{x}] & \triangleq - \int_{-\infty}^{+\infty}f(\mathbf{x})\ln f(\mathbf{x}) \mathrm{d}\mathbf{x} \\ & = -\mathbb{E}[\ln \mathcal{ N}(\mathbf{x}|\boldsymbol{\mu}, \boldsymbol{\Sigma})] \\ 
    &= \frac{1}{2}\ln \mathrm{det}(\boldsymbol{\Sigma}) + \frac{D}{2}\ln(2\pi e) \approx \ln \mathrm{det}(\boldsymbol{\Sigma}).
\end{aligned}
\end{equation}
Let $f(\mathbf{x})$ be capsule $j$'s variational posterior: $q^{\star}(\boldsymbol{\mu}_j, \boldsymbol{\Lambda}_j) = \mathcal{N}(\boldsymbol{\mu}_j|\boldsymbol{\mathrm{m}}_j, (\kappa_j \boldsymbol{\Lambda}_j)^{-1})\mathrm{Wi}(\boldsymbol{\Lambda}_j|\boldsymbol{\Psi}_{j},\nu_{j})$, where $\mathbf{m}_j$, $\kappa_j$, $\boldsymbol{\Psi}_j$ and $\nu_j$ are the updated prior parameters for a capsule $j$ as detailed in Algorithm 1. We then approximate the entropy
\begin{equation}
\begin{aligned}
     \mathbb{H}[q^{\star}&(\boldsymbol{\mu}_j, \boldsymbol{\Lambda}_j)] \approx \mathbb{E}[\ln \mathrm{det}(\boldsymbol{\Lambda}_j)] = \\ &= \sum_{i=0}^{D-1}\psi \bigg(\frac{\nu_j - i}{2}\bigg) + D\ln 2 + \ln \mathrm{det}(\boldsymbol{\Psi}_j),
\end{aligned}
\end{equation}
where $\psi(\cdot)$ is the digamma function, and we use $\mathbb{E}[\ln \mathrm{det}(\boldsymbol{\Lambda}_j)]$ to indirectly measure the differential entropy of capsule $j$'s variational posterior distribution, up to constant factors. Intuitively, the determinant of the precision matrix measures the concentration of data points across the volume defined by the matrix. The higher the concentration the higher the agreement is among votes for capsule $j$. To compute any capsule $j$'s activation probability $a_j$, we pass in both its mixing proportion and posterior entropy, as a measure of vote agreement through a logistic non-linearity
\begin{equation}
    a_j = \sigma \Big(\beta_a - \big(\beta_u + \mathbb{E}[\ln \pi_j] + \mathbb{E}[\ln \mathrm{det}(\boldsymbol{\Lambda}_j)]\big) \odot \mathbf{r}_j \Big),  
\end{equation}
where $\beta_a$ and $\beta_u$ are learnable offset parameters as in~\cite{hinton2018matrix}. Unlike EM or Dynamic routing, we only activate the capsules after the routing iterations. We find this to have a stabilising effect during training, and we can add in the expected mixing coefficients as a weight on the differential entropy of each capsule, encouraging a trade-off between activating the capsule with the most votes and our measure of how concentrated they are. This decision is in part motivated by context-dependent weighted information and entropy principles, wherein two separate low probability events incurring equally high surprisal can yield contextually unequal informative value~\cite{guiacsu1971weighted}. 

Note that the updated prior parameters $\mathbf{m}_j$, $\kappa_j$, $\boldsymbol{\Psi}_j$ and $\nu_j$, have a dependency on the routing weights $\mathbf{r}_j = \sum_i \mathbf{r}_{ij} \odot \mathbf{a}_i$, which represent the amount of data assigned to capsule $j$, weighted by the previous capsule layer activations. From the perspective of any capsule $j$'s cluster, previous layer activations $\mathbf{a}_i$ simply dictate how important each data point is. 
\subsection{Capsule-VAE}
It is possible to transform our CapsNet into a Variational Autoencoder (VAE)~\cite{kingma2013auto} by sampling from the approximate variational posterior on the capsule parameters $q^{\star}(\boldsymbol{\mu}_j, \boldsymbol{\Lambda}_j)$. We can do so by saving the updated prior parameters $\mathbf{m}_j$, $\kappa_j$, $\boldsymbol{\Psi}_j$ and $\nu_j$, at the end of the routing procedure of the final layer, and output the capsule means and precisions as latent code. Recall that the approximate posterior on the mean and precision of any capsule $j$ is a Gaussian-Wishart $q^{\star}(\boldsymbol{\mu}_j, \boldsymbol{\Lambda}_j) = \mathcal{N}(\boldsymbol{\mu}_j|\boldsymbol{\mathrm{m}}_j, (\kappa_j \boldsymbol{\Lambda}_j)^{-1})\mathrm{Wi}(\boldsymbol{\Lambda}_j|\boldsymbol{\Psi}_{j},\nu_{j})$, and we can sample from this distribution in the following way
\begin{equation}
\begin{aligned}
    \boldsymbol{\Lambda}_j \ | \ \boldsymbol{\Psi}_j, \nu_j \ &\sim  \mathrm{Wi}(\boldsymbol{\Psi}_{j},\nu_{j}) \\
    \boldsymbol{\mu}_j \ | \ \boldsymbol{\Lambda}_j, \mathbf{m}_j, \kappa_j &\sim \mathcal{N}(\boldsymbol{\mathrm{m}}_j, (\kappa_j \boldsymbol{\Lambda}_j)^{-1}).
\end{aligned}
\end{equation}
It is straight forward to condition the sample on the target class capsule during training based on the label, and make the process differentiable using the reparameterisation trick
\begin{equation}
    z \ \sim \mathcal{N}(\boldsymbol{\mu}_j, \boldsymbol{\sigma}_j) = g_{\boldsymbol{\mu}_j,\boldsymbol{\sigma}_j}(\epsilon) = \boldsymbol{\mu}_j + \boldsymbol{\epsilon} \odot \boldsymbol{\sigma}_j
\end{equation}
where $\boldsymbol{\epsilon} \sim \mathcal{N}(0, \mathbf{I})$, and $\boldsymbol{\sigma}_j$ $\triangleq$ $\mathrm{diag}(\boldsymbol{\Lambda}_j)^{- \frac{1}{2}}$. This formulation also reduces computational time since we can avoid explicit redo of VB for each sample. Capsule-VAEs are interesting models as the output latent code is composed of capsule instantiation parameters, and we know from \cite{sabour2017dynamic} that each capsule dimension learns to encode different variations of object properties that we can visualise/tweak. We leave further exploration of these ideas and analysis of Capsule-VAEs to future work.
\section{Related Work}
Capsules were first introduced by~\cite{hinton2011transforming}, wherein the encoding of instantiation parameters was established in a transforming autoencoder. More recently, work by~\cite{sabour2017dynamic} achieved state-of-the-art performance on MNIST with a shallow CapsNet, using a Dynamic routing algorithm. Shortly after, EM routing was proposed in~\cite{hinton2018matrix}, replacing capsule vectors with matrices to reduce the number of parameters. State-of-the-art performance was achieved on smallNORB, outperforming CNNs. More recently, Group Equivariant CapsNets were proposed in~\cite{lenssen2018group}, leveraging ideas from group theory to guarantee equivariance and invariance properties. In~\cite{zhang2018fast} a new routing algorithm based on kernel density estimation was proposed, providing a speed up compared to EM routing. Capsules have also been extended to action recognition in videos by~\cite{duarte2018videocapsulenet}, where the propose to average the votes before routing them for speed. Work in~\cite{zhang2018cappronet} proposes learning groups of capsule subspaces and project embedded features onto these subspaces. Despite these interesting works among others, CapsNets are still difficult to train and the original state-of-the-art benchmarks are yet to be beaten fairly.
\begin{figure*}[t]
    \centering
    \includegraphics[width=2.11\columnwidth]{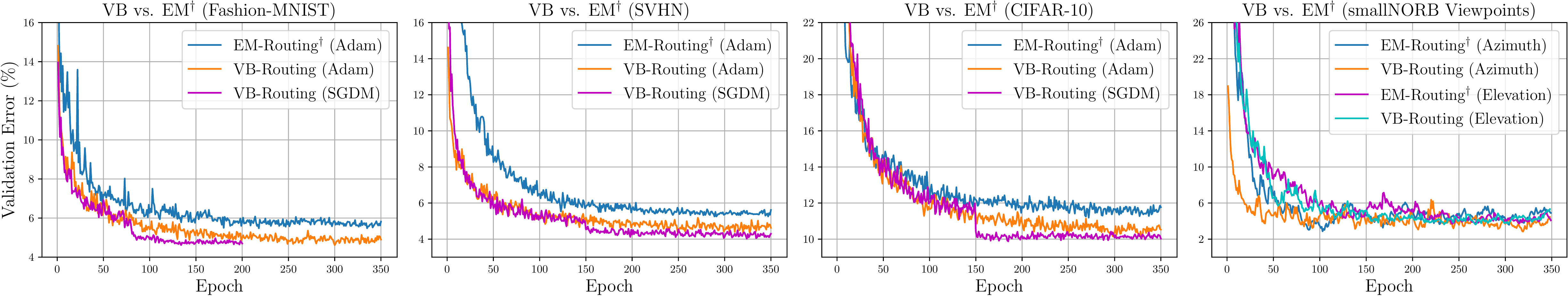}
 \caption{Direct comparison between VB and EM$^{\dagger}$ routing validation set error using identical networks and hyperparameters.} 
\label{loss_curves}
\end{figure*}
\section{Experiments}
\paragraph{Capsule Network Architecture}\label{sec: architecture}
Our CapsNet follows the EM routing formulation and comprises 4 capsule layers, starting with a primary capsule ({\fontfamily{cmtt}\selectfont PrimaryCaps}) layer followed by 3 convolutional capsule ({\fontfamily{cmtt}\selectfont ConvCaps}) layers. The stem of the network consists of a $5 \times 5$ {\fontfamily{cmtt}\selectfont Conv} layer using $\mathcal{F}$ filters and stride $2$, and is followed by two $3 \times 3$ {\fontfamily{cmtt}\selectfont Conv} layers with $\mathcal{F}$ filters each, all using {\fontfamily{cmtt}\selectfont BatchNorm} and {\fontfamily{cmtt}\selectfont ReLU} activations. The {\fontfamily{cmtt}\selectfont PrimaryCaps} layer transforms the $\mathcal{F}$ filters into $d_1$ capsule pose 4x4 matrices and $d_1$ activations using $1 \times 1$ convolutions. This is followed by a $3 \times 3$ {\fontfamily{cmtt}\selectfont ConvCaps} layer with $d_2$ capsules types and stride $2$, and a $3 \times 3$ {\fontfamily{cmtt}\selectfont ConvCaps} layer with $d_3$ capsule types and stride $1$. The final {\fontfamily{cmtt}\selectfont ConvCaps} layer shares weight matrices across spatial dimensions, yielding a capsule for each class of $d_4$ classes, and we perform coordinate addition as in~\cite{hinton2018matrix}. In summary, we describe our network architectures using the notation $\{\mathcal{F}, d_1, d_2, d_3, d_4\}$.
\paragraph{Objective Function}
We experiment with both a negative likelihood loss $\mathcal{L}_{\mathrm{NLL}}$, and the spread loss $\mathcal{L}_{\mathrm{SL}}$ in~\cite{hinton2018matrix}, then add the VAE loss $\mathcal{L}_{\mathrm{VAE}}$ as an optional capsule reconstruction based regulariser
\begin{equation}
\begin{aligned}
    \mathcal{L}_{\mathrm{SL}} &= \sum_{i \neq j} \max\big(0, m - (a_t - a_j)\big)^2, \\ 
    \mathcal{L}_{\mathrm{NLL}} &= -\sum_{j} \sum_{k} \mathbf{y}_{jk} \log\big(\widehat{\mathbf{y}}_{jk}\big).
\end{aligned}
\end{equation}
\begin{equation}
\begin{aligned}
    \mathcal{L}_{\mathrm{VAE}} = \frac{1}{2} \sum_{d} \big(\boldsymbol{\sigma}_{jd}^2 + &\boldsymbol{\mu}_{jd}^2 - \ln\boldsymbol{\sigma}_{jd}^2 - 1\big) \\ &+ \frac{1}{K}  \sum_{k}
    \norm{\mathbf{x}_{k} - f({\mathbf{x}_k})}^2_{F}.
\end{aligned}
\end{equation}
The total loss is a linear combination of a classification loss and the optional VAE loss i.e. $\mathcal{L} = \mathcal{L}_{\mathrm{NLL}} + \eta \mathcal{L}_{\mathrm{VAE}}$. CapsNet regularisation by reconstruction was first proposed in~\cite{sabour2017dynamic} with a fully-connected decoder, in our VAE we use a simple 5 layer deconvnet.
\paragraph{Uninformative Priors} We set the gaussian priors on the mean parameters $\mathbf{m}_0$ to be zeros with precision scaling $\kappa_0=1$, and the wishart priors on the precision matrix $\boldsymbol{\Psi}_0$ to be identities $\mathbf{I}_D$ with degrees of freedom $\nu_0=D+1$. For the diagonal case, $\boldsymbol{\lambda}_0$ is a vector of $1$'s. These priors have a regularising effect since they encourage the parent capsule clusters $j$ to remain close to the origin, and not to be too irregular in shape. The Dirichlet prior on the mixing coefficients $\alpha$ is set to $1$, and reducing this value favours routing solutions with less active parent capsules. In section~\ref{priors}, we provide some analysis on sensitivity to prior initialisations.
\paragraph{Weight Initialisation}
CapsNets are known to be difficult to train, in fact, the EM routing results were yet to be fairly matched before this paper. With that said, we provide some valuable suggestions for practitioners on how to initialise the various parameters of the model that worked well for us experimentally, and helped stabilise training significantly. 
We offer the following two ways of initialising the $\mathbf{W}_{ij}$ viewpoint-invariant transformation weight matrices:
\begin{enumerate}[(i)]
  \item As identities $\textbf{I}_4 \in \mathbb{R}^{4 \times 4}$ with added random uniform noise $\boldsymbol{\epsilon} \sim \mathrm{Unif}(0,b)$ on the off diagonal entries. In this way, at the start of training the capsule pose transformations don't stray too far from computing the identity function, which we find to have a stabilising effect.
  \item To help maintain constant variance of activations across capsule layers and help avoid exploding/vanishing gradients, we propose initialising $\mathbf{W}_{ij}$ with a modified~\cite{glorot2010understanding} scheme as
\begin{equation}
\begin{aligned}
    \mathbf{W}_{ij} \sim \mathrm{Unif} \big(&-r, r\big), \\
    &r = \frac{\sqrt{6}}{(c_ik^2p^2 + d_jk^2p^2)^\frac{1}{2}},
\end{aligned}
\end{equation}
where $c_i$ and $d_j$ denote the number of capsules types in layers $\mathcal{L}_i$ and $\mathcal{L}_j$, $k$ is the convolutional kernel size and $p^2$ is the number of neurons per capsule matrix ($4\times4$).
\end{enumerate}
Lastly, we also normalise the argument of the logistic function for $a_j$ using {\fontfamily{cmtt}\selectfont BatchNorm} without the learnable parameters $\gamma$ and $\beta$. This restricts the range of input values from being too high/low and helps prevent vanishing gradients. 
\subsection{Image Classification Results}
\label{results_section}
The main comparative results are reported in Table~\ref{results table}, using smallNORB~\cite{lecun2004learning}, Fashion-MNIST~\cite{xiao2017fashion}, SVHN~\cite{netzer2011reading} and CIFAR-10~\cite{krizhevsky2009learning}. In all cases, we use the diagonal parameterisation in Eq.~\eqref{eq: Gaussian-Gamma}, 3 VB routing iters and batch size 32. All hyperparameters were tuned using validation sets, then models were retrained with the full training set until convergence before testing.
\paragraph{smallNORB} 
smallNORB consists of grey-level stereo 96x96 images of 5 objects. Each object is given at 18 different azimuths (0-340), 9 elevations and 6 lighting conditions, and there are 24,300 training and test set images each. Following~\cite{hinton2018matrix}, we standardise and resize all images to 48x48 and take random 32x32 crops during training. At test time, we simply center crop the images to 32x32. Our best model $\{64, 16, 16, 16, 5\}$ was trained for 350 epochs using Adam, $\mathcal{L}_{\mathrm{NLL}}$ loss, and 3e-3 initial learning rate with exponentially decay. A 20\% validation split of the training set was used to tune hyperparameters. As reported in Table~\ref{results table}, we achieve a best test error rate of $\mathbf{1.55}$\% ($\mathbf{1.6}\%\scriptsize{\pm.06}$ over 5 runs) compared to the previous state-of-the-art $\mathbf{1.8}$\% reported in~\cite{hinton2018matrix}. Note that by averaging multiple crops at test time they can get $\mathbf{1.4}$\% and we reach $\mathbf{1.29}$\%. Our result is obtained without adding random brightness/contrast or any other augmentations/deformations during training. We also stress that our capsule network has $\simeq\textbf{50}\%$ fewer capsules. 
\begin{table}[t]
\footnotesize
\centering
\caption{Comparing novel viewpoint generalisation. ($\dagger$) denotes our implementation of EM with same network as VB.}
\label{viewpoints}
\begin{tabular}{lccc|ccc}
\toprule  
\textbf{Viewpoints} & \multicolumn{3}{c}{\textbf{Azimuth} \ (\%)} & 
\multicolumn{3}{c}{\textbf{Elevation} \ (\%)} \\
(Test) & VB & EM$^{\dagger}$ & EM & VB & EM$^{\dagger}$ & EM \\
\midrule
Novel & \textbf{11.33} & 12.67 & 13.5 & \textbf{11.59} & 12.04 & 12.3 \\  
Familiar & 3.71 & 3.72 & 3.7 & 4.32 & 4.29 & 4.3 \\ 
\bottomrule
\end{tabular}
\end{table}
\begin{figure}[t]
    \centering
	\includegraphics[width=.975\columnwidth]{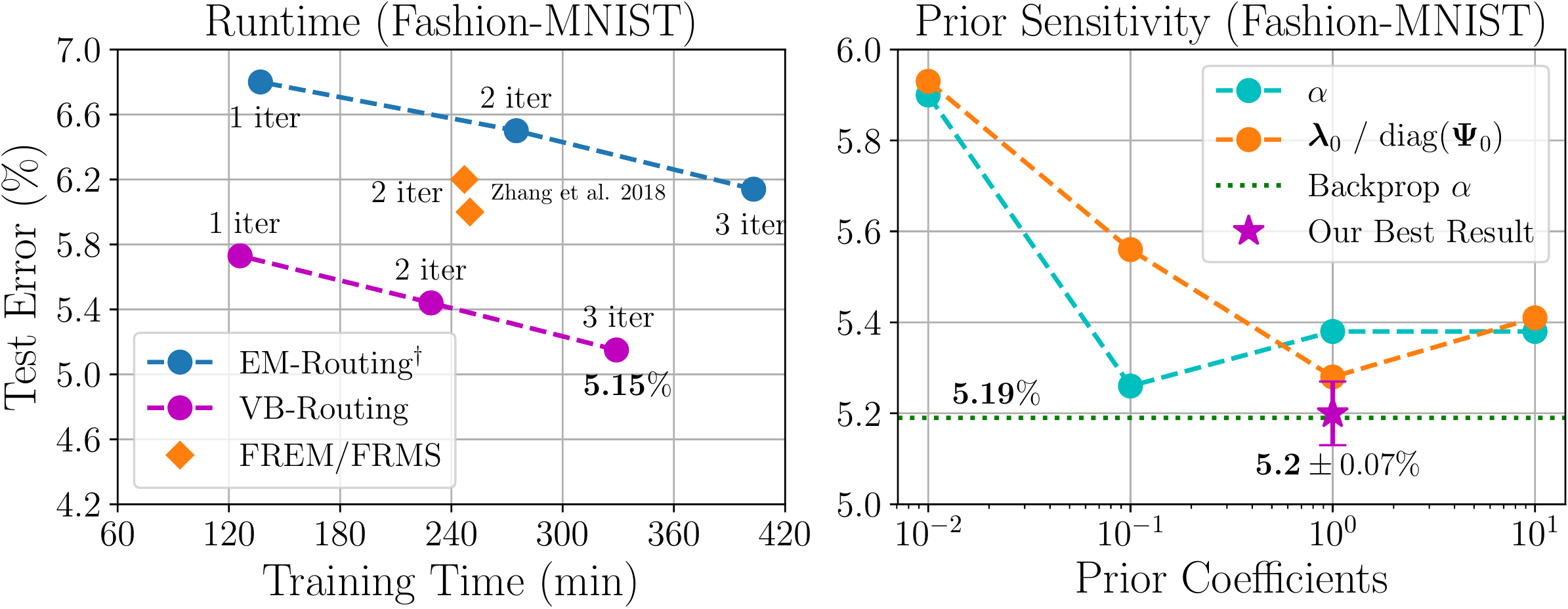}
	\caption{Test error (\%) sensitivity to priors (Right), and runtime/error comparisons using \{3,2,1\} routing iters (Left).}
	\label{runtime_sensitivity}
\end{figure}
\paragraph{Fashion-MNIST} 
Fashion-MNIST is a more difficult version of MNIST comprised of 10 clothing item classes. The images are 28x28 and the training/test sets have 60,000 and 10,000 examples respectively. We normalise and pad to 36x36, and randomly crop 32x32 image patches during training. At test time we pad the images to 32x32. Our best model $\{64, 16, 16, 16, 10\}$ was trained for 200 epochs using $\mathcal{L}_{\mathrm{NLL}}$ loss, with SGDM and a weight decay of 1e-6. The initial learning rate was set to 0.1 with step decay at 80, 120, 160 epochs and a decay rate of 0.1. As reported in Table~\ref{results table} we achieve a best test error rate of $\mathbf{5.15}$\% ($\mathbf{5.2}\%\scriptsize{\pm.07}$ over 3 runs) outperforming other works with fewer parameters. 
\paragraph{SVHN}
SVHN comprises challenging real-world 32x32 images of house numbers (10 digit classes). We trained on the core training set only, consisting of 73,257 examples and tested on the 26,032 in the test set. We normalise and pad to 40x40 and take random 32x32 crops during training. Our best model $\{64, 16, 
32, 32, 10\}$ was trained for 350 epochs using $\mathcal{L}_{\mathrm{NLL}}$ loss with SGDM. The initial learning rate was set to 0.1 with step decay at 150, 250, 300 epochs and a decay rate of 0.1. As reported in Table~\ref{results table}, we achieved a best test error of $\mathbf{3.87}$\% ($\mathbf{3.9}\%\scriptsize{\pm.06}$ over 3 runs), outperforming the Dynamic routing capsules~\cite{sabour2017dynamic} and others, with significantly fewer parameters.
\paragraph{CIFAR-10}
CIFAR-10 consists of 60,000 32x32 colour images of 10 classes. There are 50,000 training and 10,000 test images. We normalise and pad to 40x40, and randomly crop 32x32 patches during training. We also apply random horizontal flips with probability $\frac{1}{2}$. Our best model $\{64, 16, 32, 32, 10\}$ was trained for 350 epochs using $\mathcal{L}_{\mathrm{NLL}}$ loss with SGDM. Initial learning rate was 0.1 with step decay at 150, 250, 300 epochs and decay rate of 0.1. We achieved a best test error of $\mathbf{11.14}$\% ($\mathbf{11.2}\%\scriptsize{\pm.09}$ over 3 runs), which is lower than EM routing~\cite{hinton2018matrix}, and using considerably fewer parameters than other capsule works (Table~\ref{results table}). CIFAR-10 is the most challenging of the 4 datasets, and to get better performance, a deeper network is required for learning better representations. To test this hypothesis, we simply replaced the stem of our capsule network with 4 residual blocks (8 layers), and achieved a much lower test error rate of $\mathbf{7.8}\%$, outperforming even deeper Residual Networks~\cite{he2016deep}.
\begin{figure}[t]
    \centering
    \bottominset{
    \resizebox{.465\columnwidth}{!}{
        \begin{tabular}{lcc}
                & \multicolumn{2}{c}{Test Accuracy (\%)} \\
                \textbf{Method} & \textbf{MNIST} & \textbf{affNIST} \\
                \midrule
                Dynamic~\scriptsize{(Sabour et al. \citeyear{sabour2017dynamic})} & \textbf{99.2} & 79 \\ 
                HitNet~\scriptsize{(Deliège et al. \citeyear{deliege2019effective})} & 99.6 & 83.03 \\
                Baseline CNN~\scriptsize{(Hinton et al. \citeyear{hinton2018matrix})} &\textbf{ 99.2} & 85.9 \\ 
                LadderCaps~\scriptsize{(Jeong et al. \citeyear{jeong2019ladder})} & 99.3 & 87.8 \\ 
                GCaps~\scriptsize{(Lenssen et al. \citeyear{lenssen2018group})} & 98.42 & 89.1 \\ 
                SparseCaps~\scriptsize{(Rawlinson et al. \citeyear{rawlinson2018sparse})} & 99$^{\ddagger}$ & 90.1 \\ 
                Attn-Routing~\scriptsize{\cite{choi2019attention}} & 99.46 & 91.6 \\ 
                SCAE~\scriptsize{\cite{kosiorek2019stacked}}$^{\ddagger}$ & 98.5 & 92.2 \\ 
                EM-Routing~\scriptsize{(Hinton et al. \citeyear{hinton2018matrix})} & \textbf{99.2} & 93.1 \\ 
                \midrule
                \textbf{VB-Routing}: $\{64,16,16,16,10\}$ & \textbf{99.2} & 96.9 \\  
                \textbf{VB-Routing}: $\{64,16,16,16,10\}$ & \textbf{99.7} & \textbf{98.1} \\
    	    \end{tabular}}
    }{\includegraphics[width=.99\columnwidth]{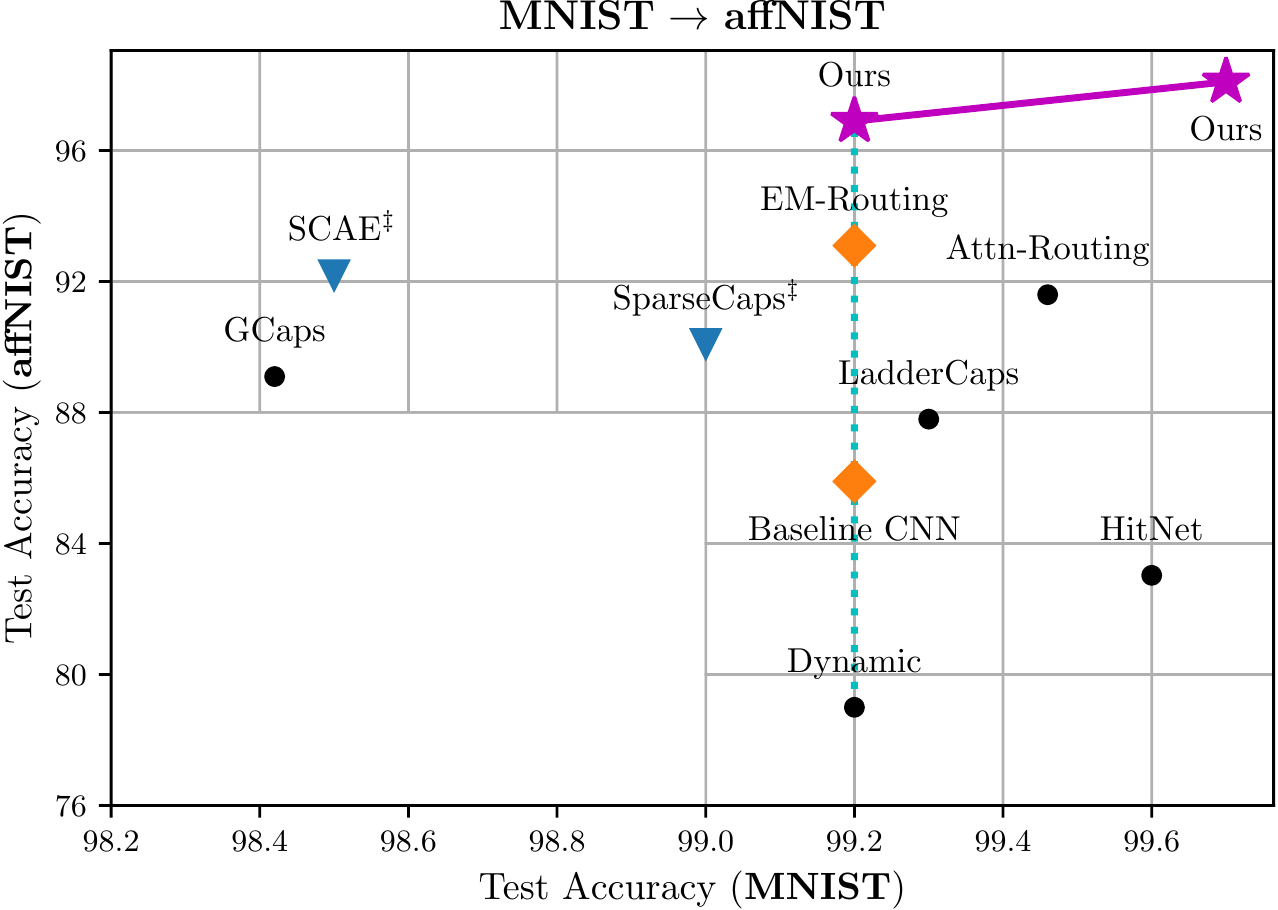}
    }{.085\columnwidth}{-.187\columnwidth}
	\caption{MNIST to affNIST generalisation performance comparisons. ($\ddagger$) denotes unsupervised learning was used, and the light blue line denotes matched performance on MNIST test set before testing on affNIST for fairness.}
    \label{affnist}
\end{figure}
\subsection{Generalisation to Novel Viewpoints}
In order to verify that our proposed capsule routing algorithm preserves generalisation to novel viewpoints, we trained our $\{64, 16, 16, 16, 5\}$ model on the smallNORB training data containing azimuths of (300, 320, 340, 0, 20, 40), and tested on the test data containing azimuths from 60 to 280. For elevation viewpoints, we trained on the 3 smaller and tested on the 6 larger elevations. During training, we validated using the portion of test data containing the same viewpoints as in training and measured the generalisation to novel viewpoints after matching the performance on familiar ones. As reported in Table~\ref{viewpoints}, we compare VB routing to the original EM routing performance in~\cite{hinton2018matrix} as well as our implementation of EM using the same network for fairness. In our experiments, VB routing does not sacrifice the ability to generalise to novel viewpoints, and outperforms EM routing in all cases.
\subsection{Affine Transformation Robustness}
To further demonstrate our method’s generalisation and invariance to affine-transformations, we train our \{64, 16, 16, 16, 10\} CapsNet on MNIST, and assess generalisation performance on the affNIST test set. AffNIST images are 40x40 so we train by randomly padding MNIST training set images as done in works we compare to. We achieve a significantly superior generalisation accuracy of $\mathbf{98.1\%}$ comparatively (Figure~\ref{affnist}). For fairer comparisons, we also match the $\mathbf{99.2\%}$ test set accuracy on MNIST reported in Dynamic/EM routing, before testing on the affNIST test set, achieving $\mathbf{96.9\%}$.
\subsection{Sensitivity to Prior Hyperparameters}
\label{priors}
We took our $\{64, 16, 16, 16, 5\}$ CapsNet, and performed sensitivity analysis on the hyperparameters of the Wishart and Dirichlet priors, with respect to test error on Fashion-MNIST (Figure~\ref{runtime_sensitivity}). We initialise $\boldsymbol{\lambda}_0 \equiv \mathrm{diag}(\boldsymbol{\Psi}_0)$ as identities scaled by coefficients $\{0.01, 0.1, 1, 10\}$. The same coefficients were used for initialising the Dirichlet prior parameter $\alpha$. In general, we find that our models are quite robust to prior initialisations in terms of final test set performance, whereas convergence speed is mildly affected. It is also possible to learn prior parameters from data via backpropagation (à la empirical Bayes), avoiding manual tuning altogether. We tested this on the Dirichlet $\alpha$ and observed no performance degradation ($\mathbf{5.19\%}$ compared to $\mathbf{5.2}\% \scriptsize{\pm 0.07}$).
\subsection{VB vs. EM Routing}
For direct comparisons with the leading capsule routing algorithm, we took our best performing models for each dataset and replaced VB with our implementation of EM. Table~\ref{results table} and Figure~\ref{loss_curves} report VB outperforming EM in terms of convergence rate, stability, and final test error with identical networks. VB routing is also almost $\mathbf{20}\%$ faster than EM. This is partly because capsule priors don't require gradient updates, and mainly because we propose to measure agreement/activate capsules after the routing iterations. As shown in Figure~\ref{runtime_sensitivity}, our method compares favourably, and we find that the number of VB routing iterations has a bigger impact on training time than test error, so we can reduce the number iterations to train faster, and still perform competitively.
\section{Conclusion}
In this paper, we propose a new capsule routing algorithm for learning a mixture of transforming gaussians via Variational Bayes. We model uncertainty over the capsule parameters in addition to the routing coefficients, which provides: (\lowerromannumeral{1}) more flexible control over capsule complexity by tuning priors to induce sparsity, and (\lowerromannumeral{2}) reduces the well known \textit{variance-collapse} problem inherent to MLE based mixture models, such as EM. We outperform the state-of-the-art on smallNORB using $\simeq$50\% fewer capsules than previously reported, achieve highly competitive performances on CIFAR-10, Fashion-MNIST, SVHN, and demonstrate significant improvement in MNIST to affNIST generalisation over previous methods. For future work, we plan to extend our Bayesian framework to obtain calibrated uncertainty estimates over predictions using capsule networks.
{\fontsize{9.0pt}{10.0pt}\selectfont
\bibliography{bibfile}}

\begin{thebibliography}{}

\bibitem[\protect\citeauthoryear{Bishop}{2006}]{bishop2006pattern}
Bishop, C.~M.
\newblock 2006.
\newblock {\em Pattern recognition and machine learning}.
\newblock springer.

\bibitem[\protect\citeauthoryear{Choi \bgroup et al\mbox.\egroup
  }{2019}]{choi2019attention}
Choi, J.; Seo, H.; Im, S.; and Kang, M.
\newblock 2019.
\newblock Attention routing between capsules.
\newblock In {\em Proceedings of the IEEE International Conference on Computer
  Vision Workshops},  0--0.

\bibitem[\protect\citeauthoryear{Deli{\`e}ge, Cioppa, and
  Van~Droogenbroeck}{2019}]{deliege2019effective}
Deli{\`e}ge, A.; Cioppa, A.; and Van~Droogenbroeck, M.
\newblock 2019.
\newblock An effective hit-or-miss layer favoring feature interpretation as
  learned prototypes deformations.
\newblock In {\em Thirty-Third AAAI Conference on Artificial Intelligence}.

\bibitem[\protect\citeauthoryear{Duarte, Rawat, and
  Shah}{2018}]{duarte2018videocapsulenet}
Duarte, K.; Rawat, Y.; and Shah, M.
\newblock 2018.
\newblock Videocapsulenet: A simplified network for action detection.
\newblock In {\em Advances in Neural Information Processing Systems},
  7610--7619.

\bibitem[\protect\citeauthoryear{Glorot and
  Bengio}{2010}]{glorot2010understanding}
Glorot, X., and Bengio, Y.
\newblock 2010.
\newblock Understanding the difficulty of training deep feedforward neural
  networks.
\newblock In {\em Proceedings of the thirteenth international conference on
  artificial intelligence and statistics},  249--256.

\bibitem[\protect\citeauthoryear{Guia{\c{s}}u}{1971}]{guiacsu1971weighted}
Guia{\c{s}}u, S.
\newblock 1971.
\newblock Weighted entropy.
\newblock {\em Reports on Mathematical Physics} 2(3):165--179.

\bibitem[\protect\citeauthoryear{He \bgroup et al\mbox.\egroup
  }{2016}]{he2016deep}
He, K.; Zhang, X.; Ren, S.; and Sun, J.
\newblock 2016.
\newblock Deep residual learning for image recognition.
\newblock In {\em Proceedings of the IEEE conference on computer vision and
  pattern recognition},  770--778.

\bibitem[\protect\citeauthoryear{Hinton, Krizhevsky, and
  Wang}{2011}]{hinton2011transforming}
Hinton, G.~E.; Krizhevsky, A.; and Wang, S.~D.
\newblock 2011.
\newblock Transforming auto-encoders.
\newblock In {\em International Conference on Artificial Neural Networks},
  44--51.
\newblock Springer.

\bibitem[\protect\citeauthoryear{Hinton, Sabour, and
  Frosst}{2018}]{hinton2018matrix}
Hinton, G.~E.; Sabour, S.; and Frosst, N.
\newblock 2018.
\newblock Matrix capsules with em routing.
\newblock In {\em International Conference on Learning Representations (ICLR)}.

\bibitem[\protect\citeauthoryear{Jeong, Lee, and Kim}{2019}]{jeong2019ladder}
Jeong, T.; Lee, Y.; and Kim, H.
\newblock 2019.
\newblock Ladder capsule network.
\newblock In {\em International Conference on Machine Learning},  3071--3079.

\bibitem[\protect\citeauthoryear{Killian \bgroup et al\mbox.\egroup
  }{2019}]{killian2019kernelized}
Killian, T.; Goodwin, J.; Brown, O.; and Son, S.-H.
\newblock 2019.
\newblock Kernelized capsule networks.
\newblock {\em arXiv preprint arXiv:1906.03164}.

\bibitem[\protect\citeauthoryear{Kingma and Welling}{2013}]{kingma2013auto}
Kingma, D.~P., and Welling, M.
\newblock 2013.
\newblock Auto-encoding variational bayes.
\newblock {\em arXiv preprint arXiv:1312.6114}.

\bibitem[\protect\citeauthoryear{Kosiorek \bgroup et al\mbox.\egroup
  }{2019}]{kosiorek2019stacked}
Kosiorek, A.~R.; Sabour, S.; Teh, Y.~W.; and Hinton, G.~E.
\newblock 2019.
\newblock Stacked capsule autoencoders.
\newblock {\em arXiv preprint arXiv:1906.06818}.

\bibitem[\protect\citeauthoryear{Krizhevsky, Hinton, and
  others}{2009}]{krizhevsky2009learning}
Krizhevsky, A.; Hinton, G.; et~al.
\newblock 2009.
\newblock Learning multiple layers of features from tiny images.
\newblock Technical report, Citeseer.

\bibitem[\protect\citeauthoryear{LeCun \bgroup et al\mbox.\egroup
  }{2004}]{lecun2004learning}
LeCun, Y.; Huang, F.~J.; Bottou, L.; et~al.
\newblock 2004.
\newblock Learning methods for generic object recognition with invariance to
  pose and lighting.
\newblock In {\em CVPR (2)},  97--104.
\newblock Citeseer.

\bibitem[\protect\citeauthoryear{Lenssen, Fey, and
  Libuschewski}{2018}]{lenssen2018group}
Lenssen, J.~E.; Fey, M.; and Libuschewski, P.
\newblock 2018.
\newblock Group equivariant capsule networks.
\newblock In {\em Advances in Neural Information Processing Systems},
  8844--8853.

\bibitem[\protect\citeauthoryear{Nair, Doshi, and
  Keselj}{2018}]{nair2018pushing}
Nair, P.; Doshi, R.; and Keselj, S.
\newblock 2018.
\newblock Pushing the limits of capsule networks.
\newblock {\em Technical note}.

\bibitem[\protect\citeauthoryear{Netzer \bgroup et al\mbox.\egroup
  }{2011}]{netzer2011reading}
Netzer, Y.; Wang, T.; Coates, A.; Bissacco, A.; Wu, B.; and Ng, A.~Y.
\newblock 2011.
\newblock Reading digits in natural images with unsupervised feature learning.

\bibitem[\protect\citeauthoryear{Phaye \bgroup et al\mbox.\egroup
  }{2018}]{phaye2018dense}
Phaye, S. S.~R.; Sikka, A.; Dhall, A.; and Bathula, D.
\newblock 2018.
\newblock Dense and diverse capsule networks: Making the capsules learn better.
\newblock {\em arXiv preprint arXiv:1805.04001}.

\bibitem[\protect\citeauthoryear{Rawlinson, Ahmed, and
  Kowadlo}{2018}]{rawlinson2018sparse}
Rawlinson, D.; Ahmed, A.; and Kowadlo, G.
\newblock 2018.
\newblock Sparse unsupervised capsules generalize better.
\newblock {\em arXiv preprint arXiv:1804.06094}.

\bibitem[\protect\citeauthoryear{Sabour, Frosst, and
  Hinton}{2017}]{sabour2017dynamic}
Sabour, S.; Frosst, N.; and Hinton, G.~E.
\newblock 2017.
\newblock Dynamic routing between capsules.
\newblock In {\em Advances in Neural Information Processing Systems (NIPS)},
  3856--3866.

\bibitem[\protect\citeauthoryear{Xiang \bgroup et al\mbox.\egroup
  }{2018}]{xiang2018ms}
Xiang, C.; Zhang, L.; Tang, Y.; Zou, W.; and Xu, C.
\newblock 2018.
\newblock Ms-capsnet: A novel multi-scale capsule network.
\newblock {\em IEEE Signal Processing Letters} 25(12):1850--1854.

\bibitem[\protect\citeauthoryear{Xiao, Rasul, and
  Vollgraf}{2017}]{xiao2017fashion}
Xiao, H.; Rasul, K.; and Vollgraf, R.
\newblock 2017.
\newblock Fashion-mnist: a novel image dataset for benchmarking machine
  learning algorithms.
\newblock {\em arXiv preprint arXiv:1708.07747}.

\bibitem[\protect\citeauthoryear{Zhang, Edraki, and
  Qi}{2018}]{zhang2018cappronet}
Zhang, L.; Edraki, M.; and Qi, G.-J.
\newblock 2018.
\newblock Cappronet: Deep feature learning via orthogonal projections onto
  capsule subspaces.
\newblock In {\em Advances in Neural Information Processing Systems},
  5814--5823.

\bibitem[\protect\citeauthoryear{Zhang, Zhou, and Wu}{2018}]{zhang2018fast}
Zhang, S.; Zhou, Q.; and Wu, X.
\newblock 2018.
\newblock Fast dynamic routing based on weighted kernel density estimation.
\newblock In {\em International Symposium on Artificial Intelligence and
  Robotics},  301--309.
\newblock Springer.

\bibitem[\protect\citeauthoryear{Zhao \bgroup et al\mbox.\egroup
  }{2019}]{zhao2019capsule}
Zhao, Z.; Kleinhans, A.; Sandhu, G.; Patel, I.; and Unnikrishnan, K.
\newblock 2019.
\newblock Capsule networks with max-min normalization.
\newblock {\em arXiv preprint arXiv:1903.09662}.

\end{thebibliography}
\bibliographystyle{aaai}
\begin{figure*}[b]
\settoheight{\tempdima}{\includegraphics[width=.19\linewidth]{example-image-a}}%
\centering\Large{\textbf{Airplane}}\par\bigskip
\centering\hspace*{-2em}\begin{tabular}{cccccc}
&\hspace{-1em} airplane &\hspace{-1.4em} animal &\hspace{-1.4em} car &\hspace{-1.4em} human &\hspace{-1.4em} truck\\
\rowname{iter 1}&\hspace{-1em}
\includegraphics[trim={0 0 0 50},clip,width=0.19\textwidth]{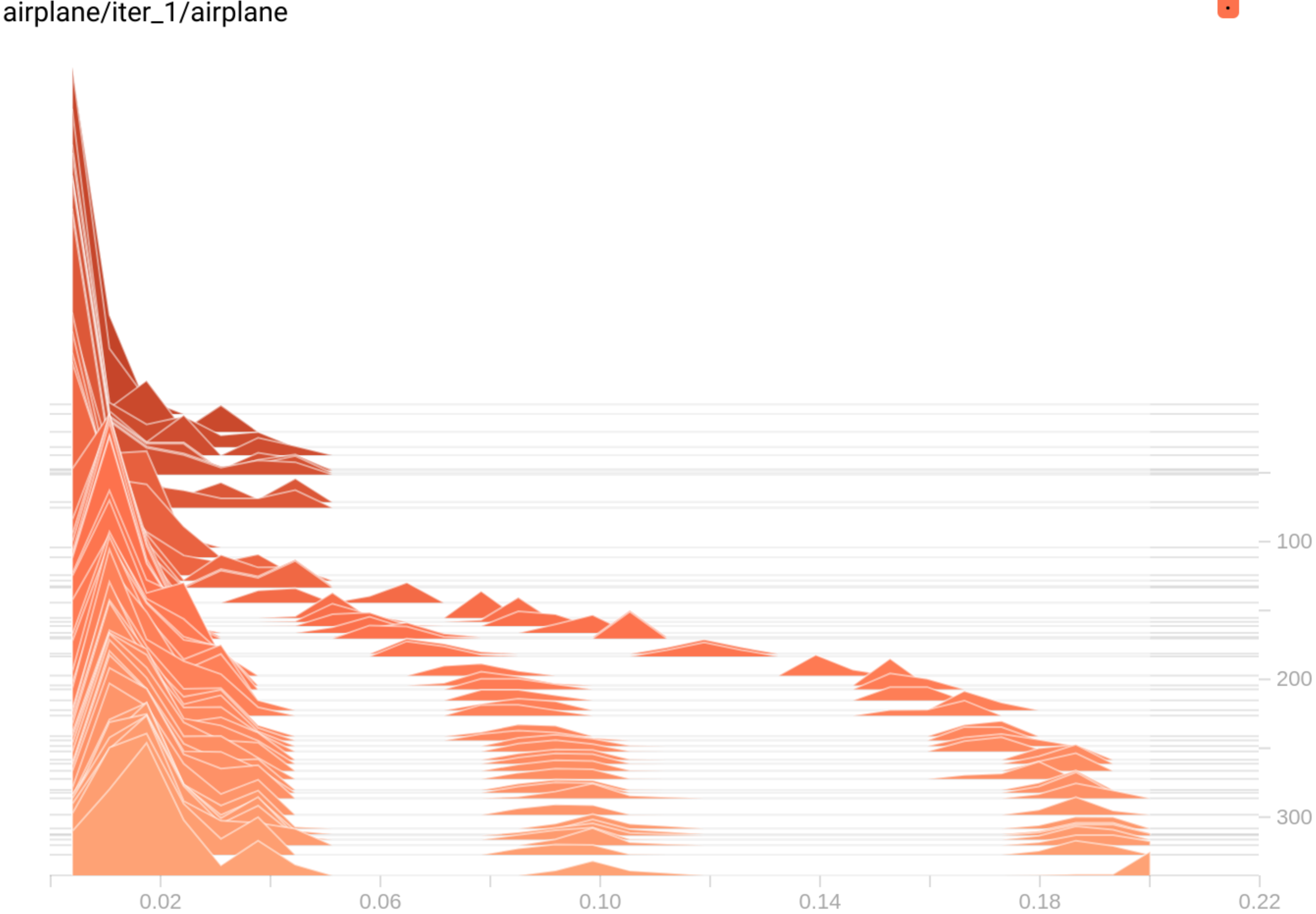}&\hspace{-1.4em}
\includegraphics[trim={0 0 0 50},clip,width=0.19\textwidth]{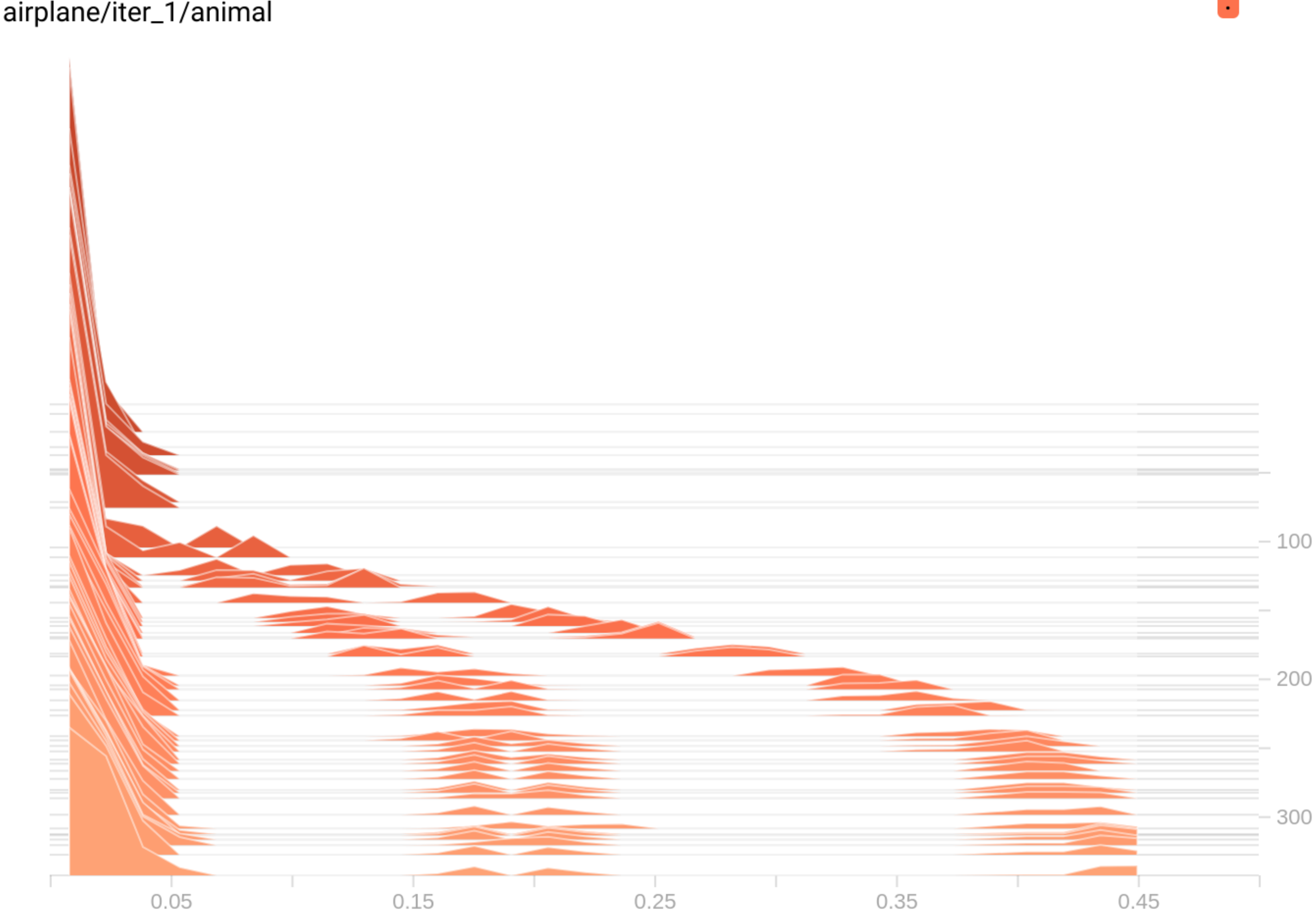}&\hspace{-1.4em}
\includegraphics[trim={0 0 0 50},clip,width=0.19\textwidth]{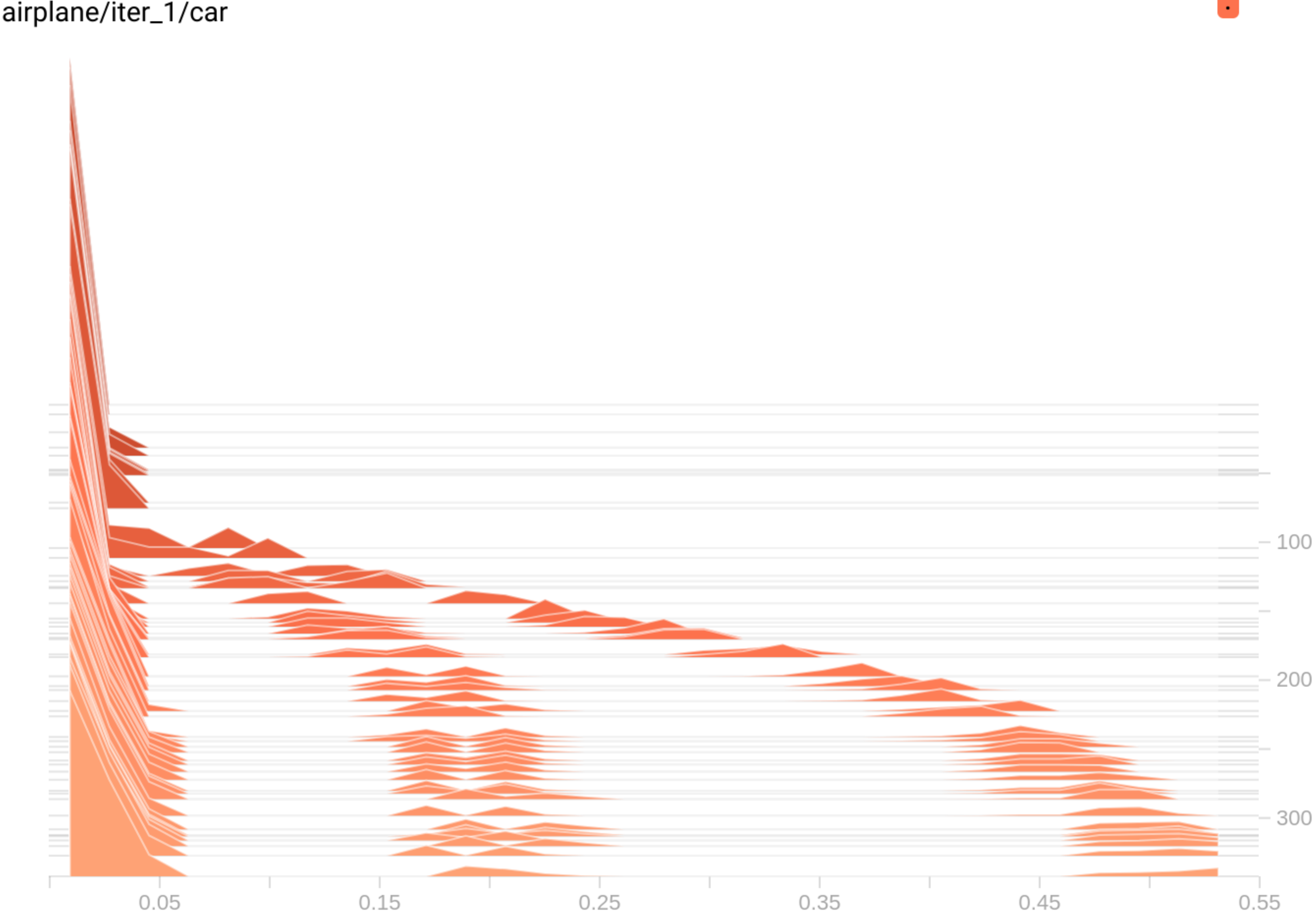}&\hspace{-1.4em}
\includegraphics[trim={0 0 0 50},clip,width=0.19\textwidth]{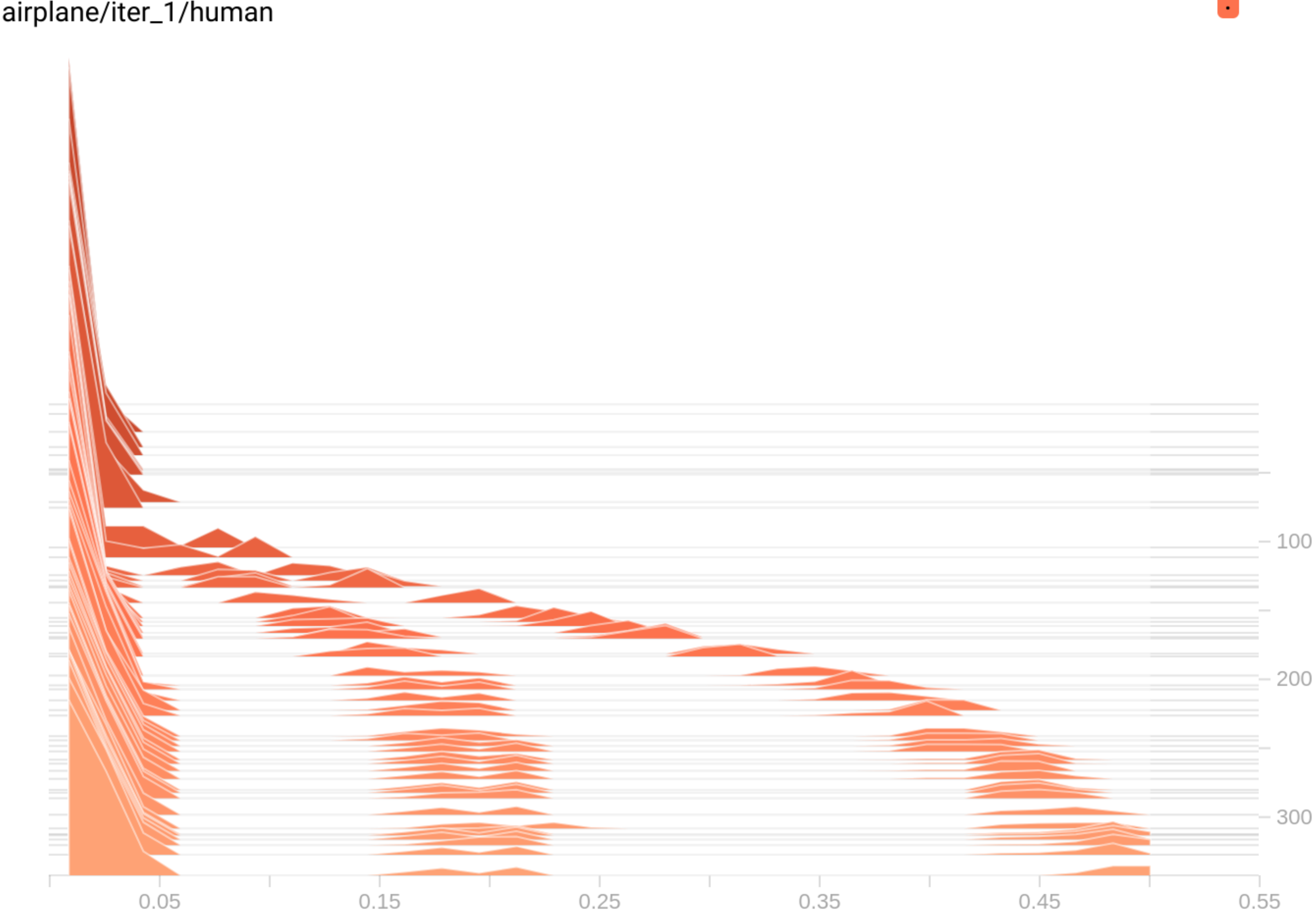}&\hspace{-1.4em}
\includegraphics[trim={0 0 0 50},clip,width=0.19\textwidth]{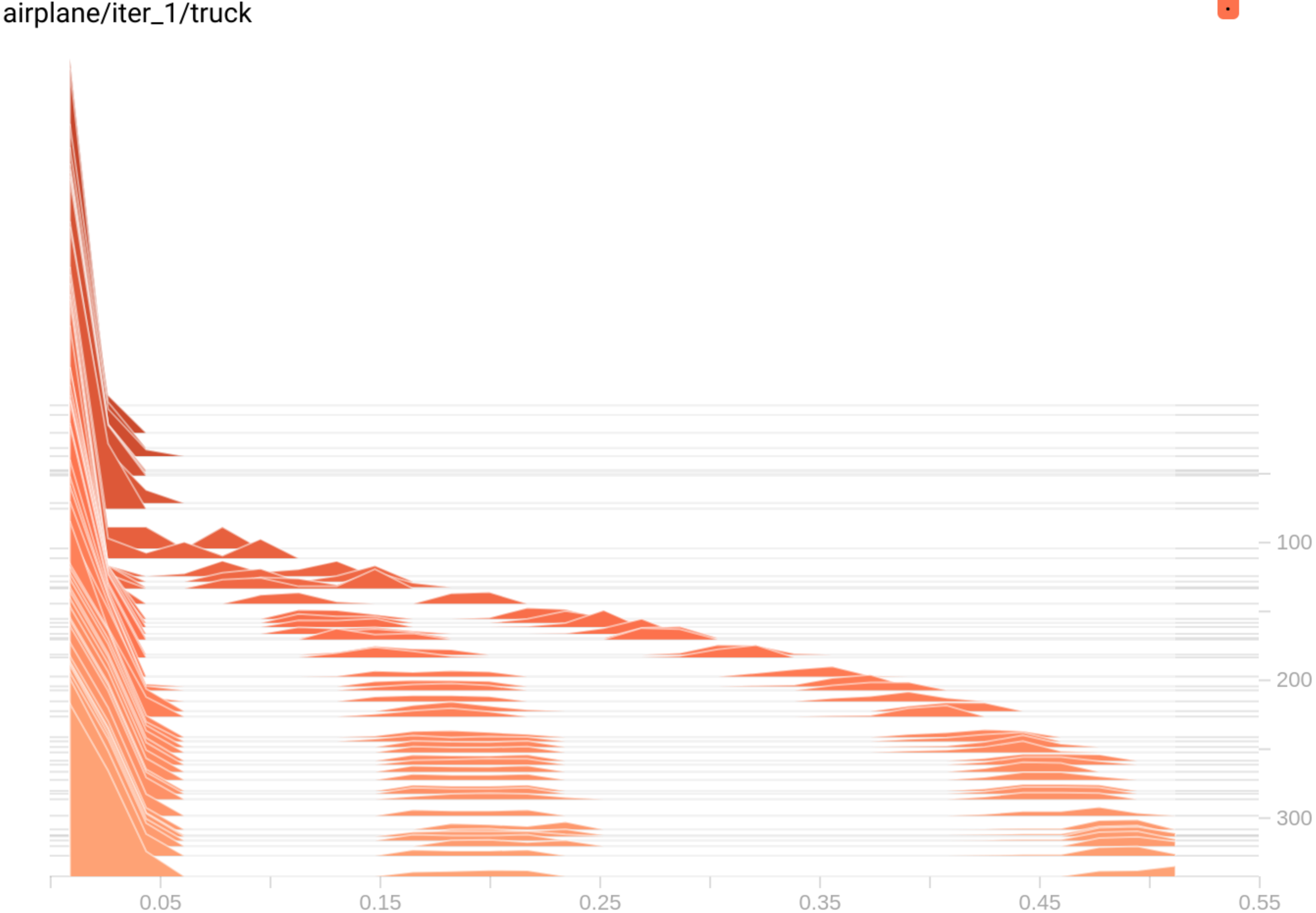} \vspace{-.4em} \\
\rowname{iter 2}&\hspace{-1em}
\includegraphics[trim={0 0 0 50},clip,width=0.19\textwidth]{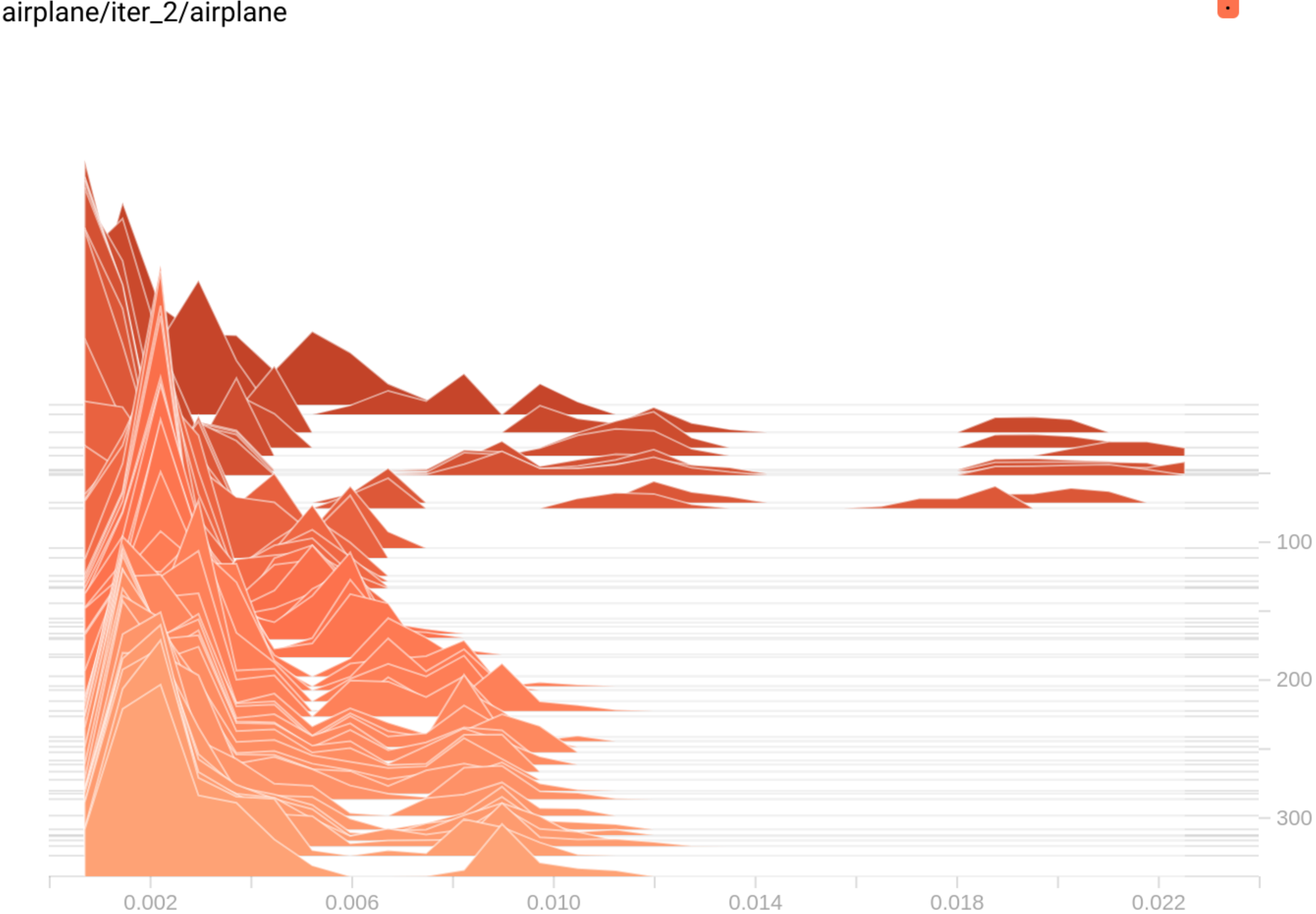}&\hspace{-1.4em}
\includegraphics[trim={0 0 0 50},clip,width=0.19\textwidth]{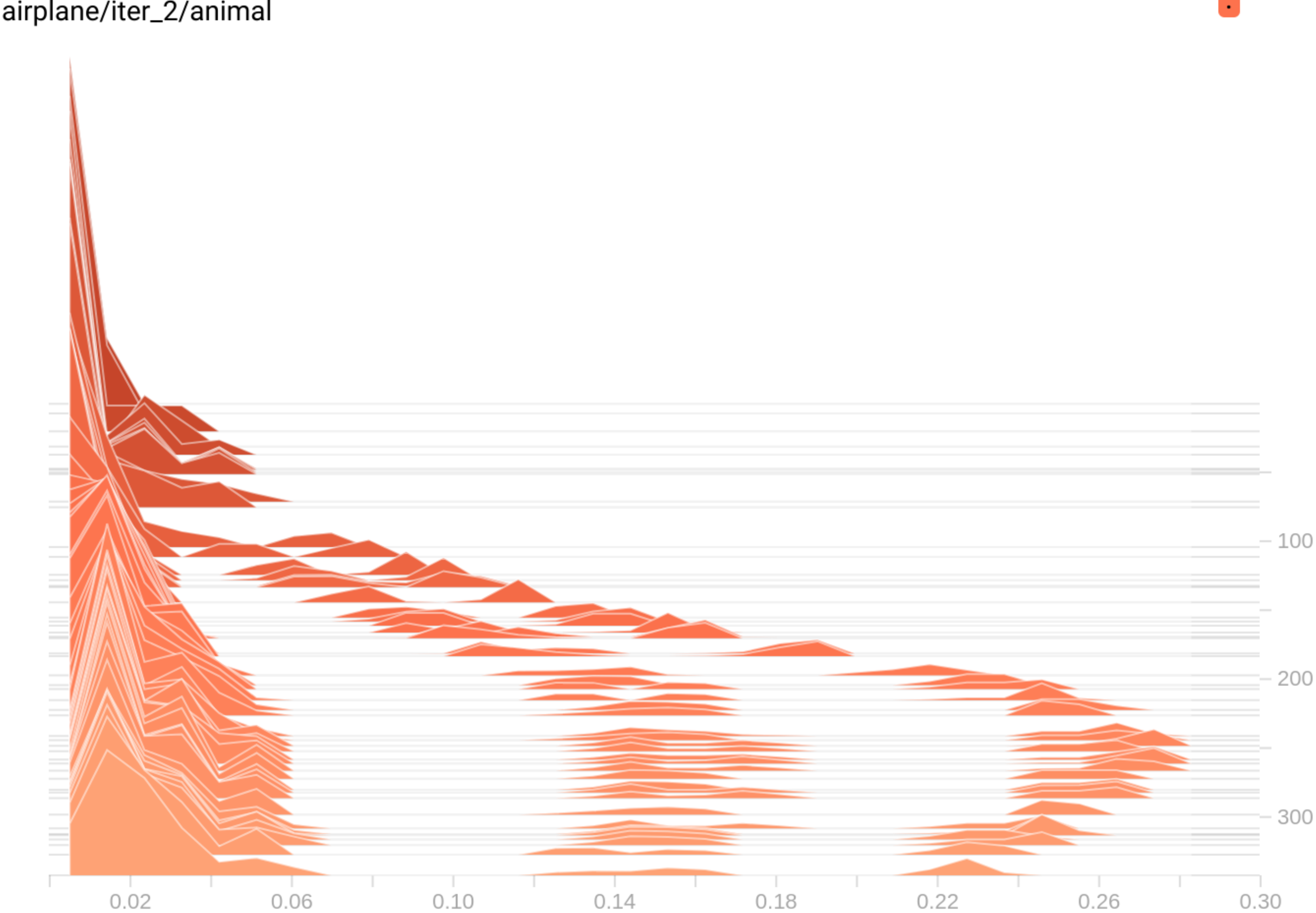}&\hspace{-1.4em}
\includegraphics[trim={0 0 0 50},clip,width=0.19\textwidth]{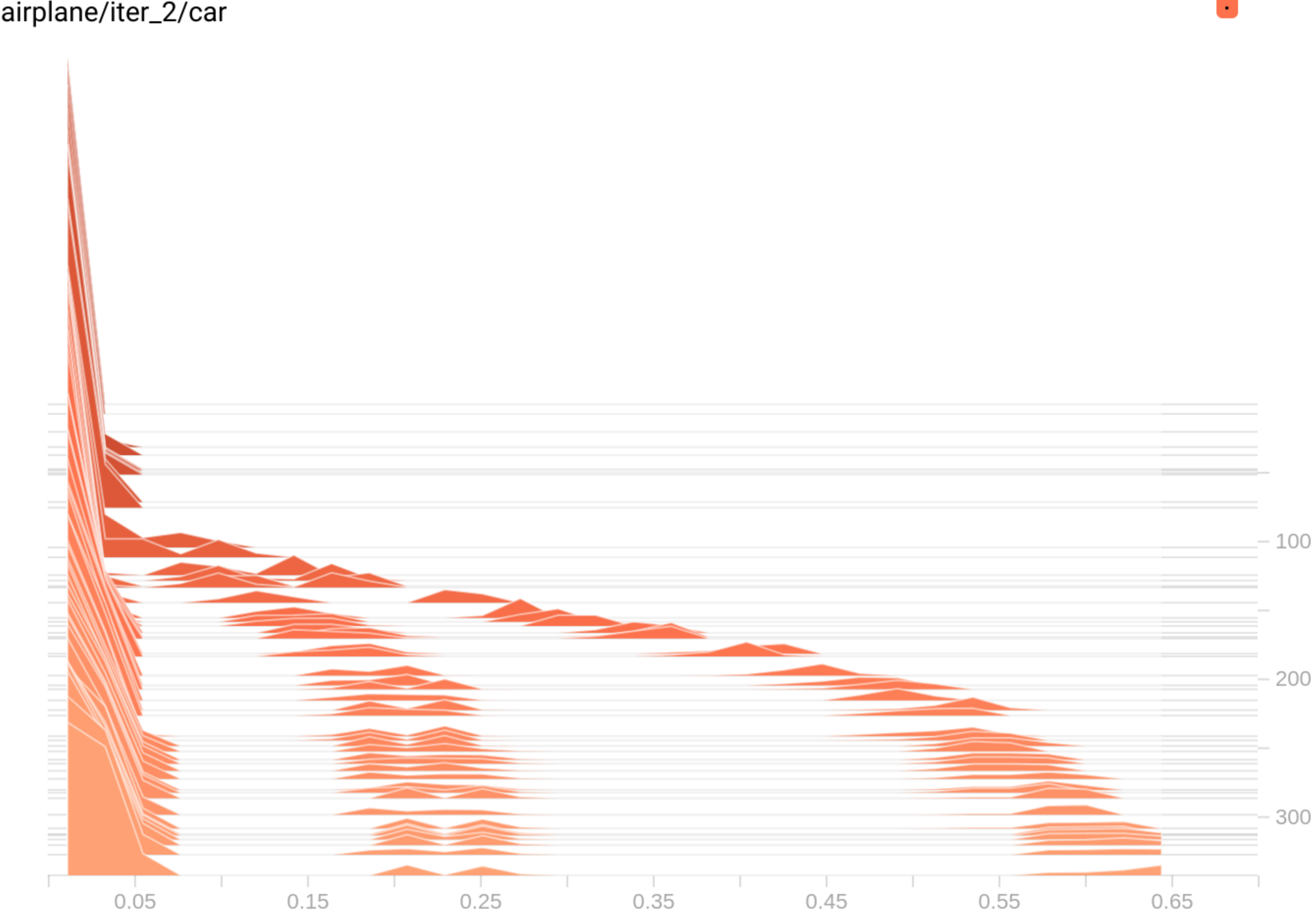}&\hspace{-1.4em}
\includegraphics[trim={0 0 0 50},clip,width=0.19\textwidth]{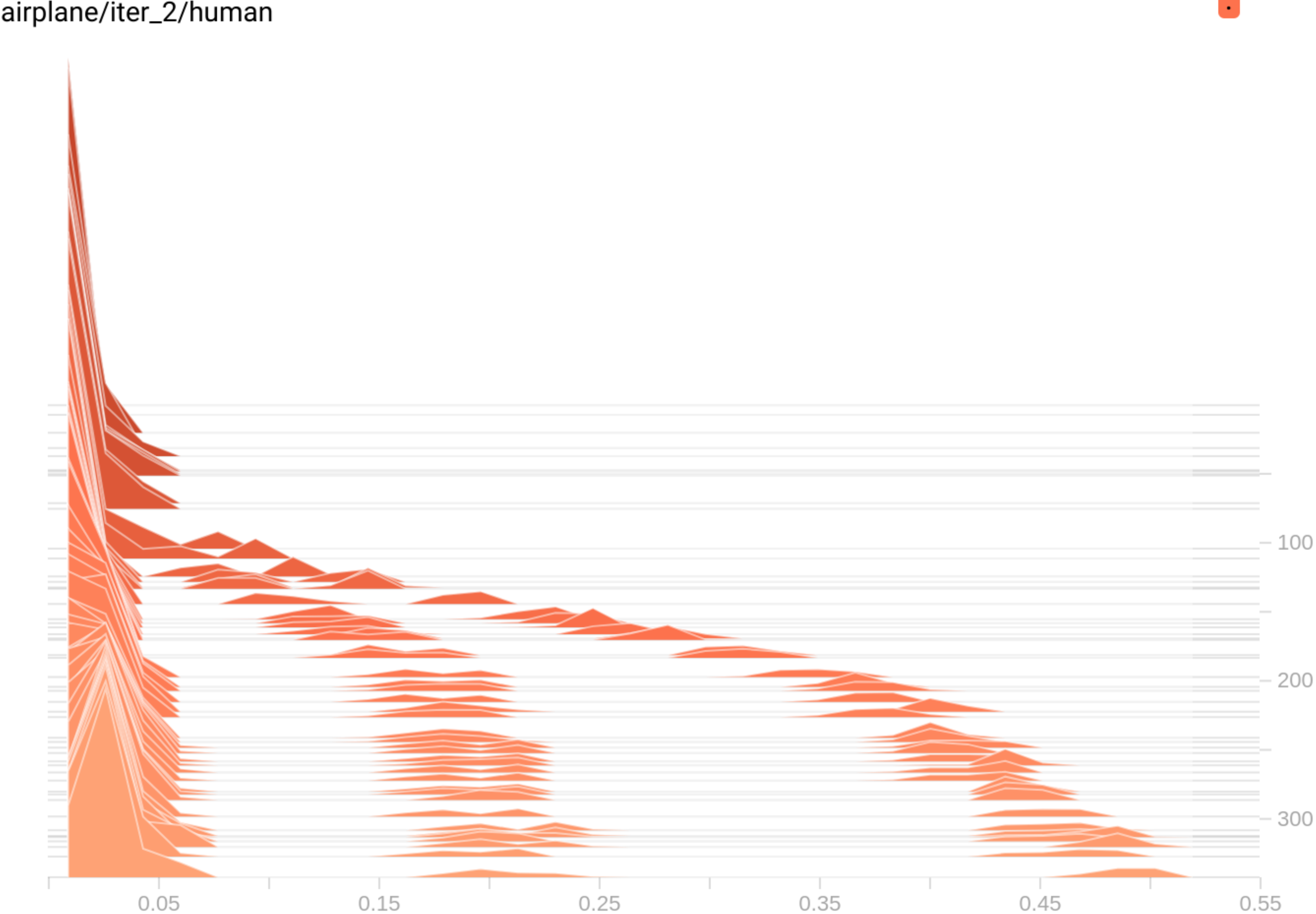}&\hspace{-1.4em}
\includegraphics[trim={0 0 0 50},clip,width=0.19\textwidth]{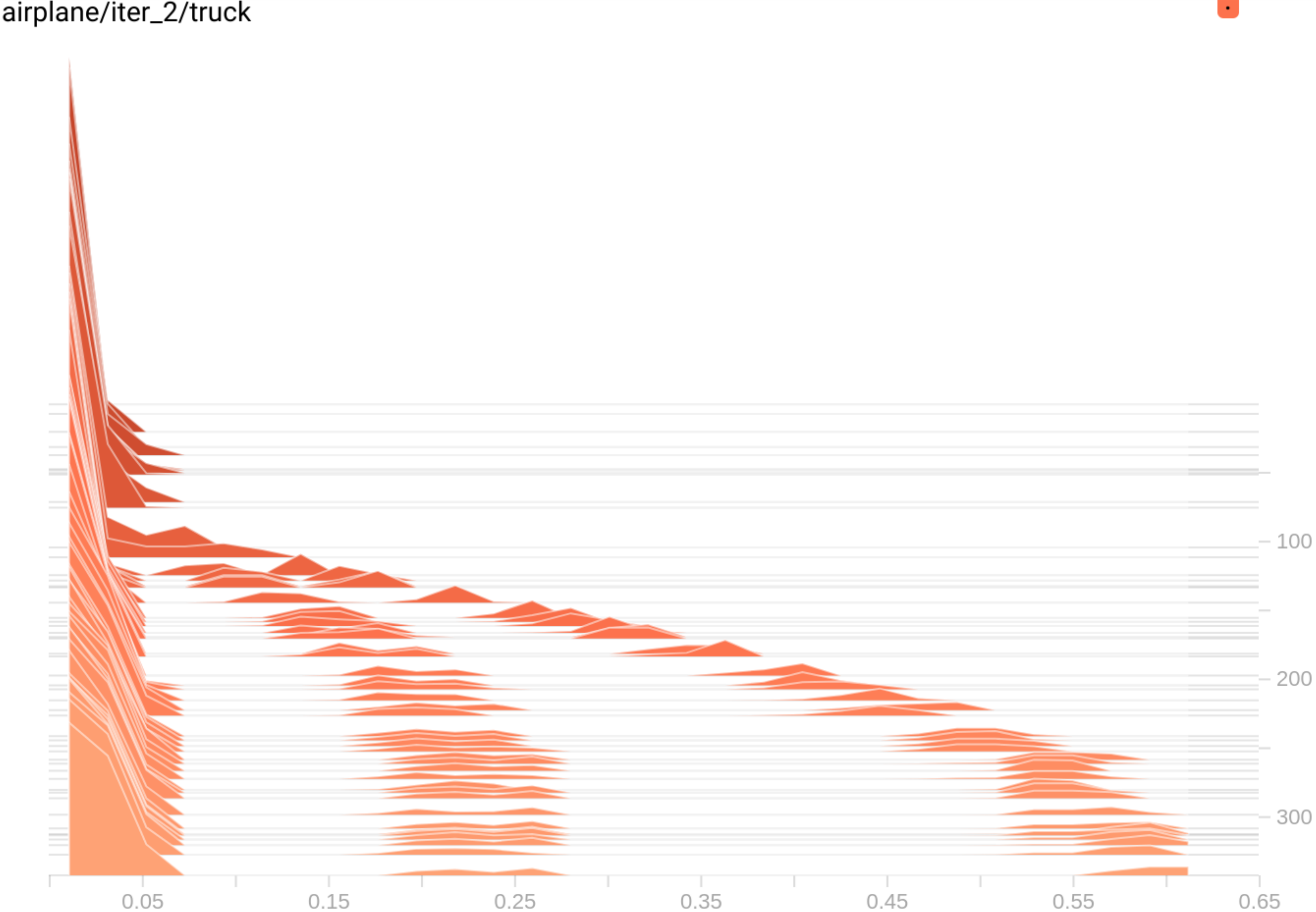} \vspace{-.4em} \\
\rowname{iter 3}&\hspace{-1em}
\includegraphics[trim={0 0 0 50},clip,width=0.19\textwidth]{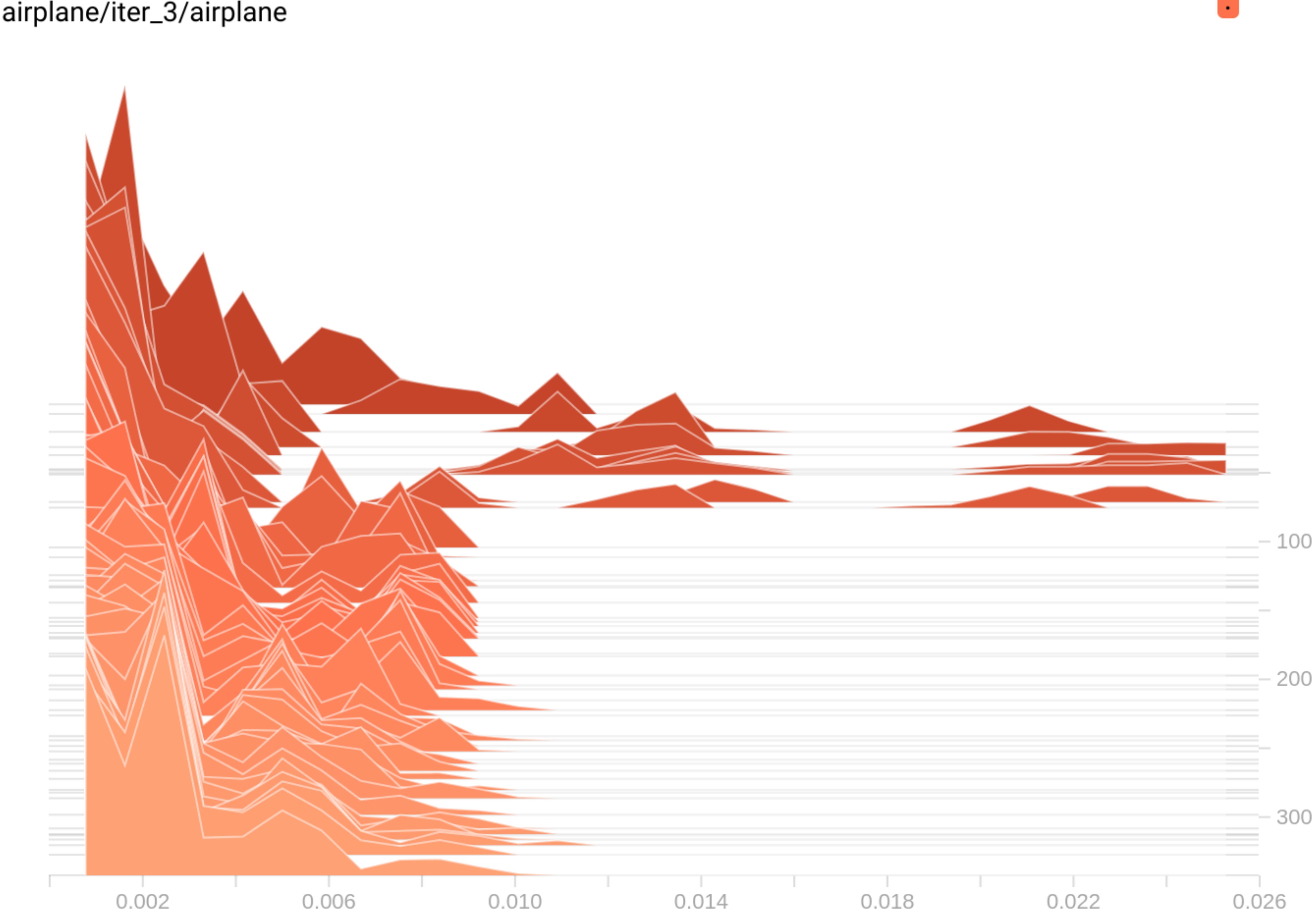}&\hspace{-1.4em}
\includegraphics[trim={0 0 0 50},clip,width=0.19\textwidth]{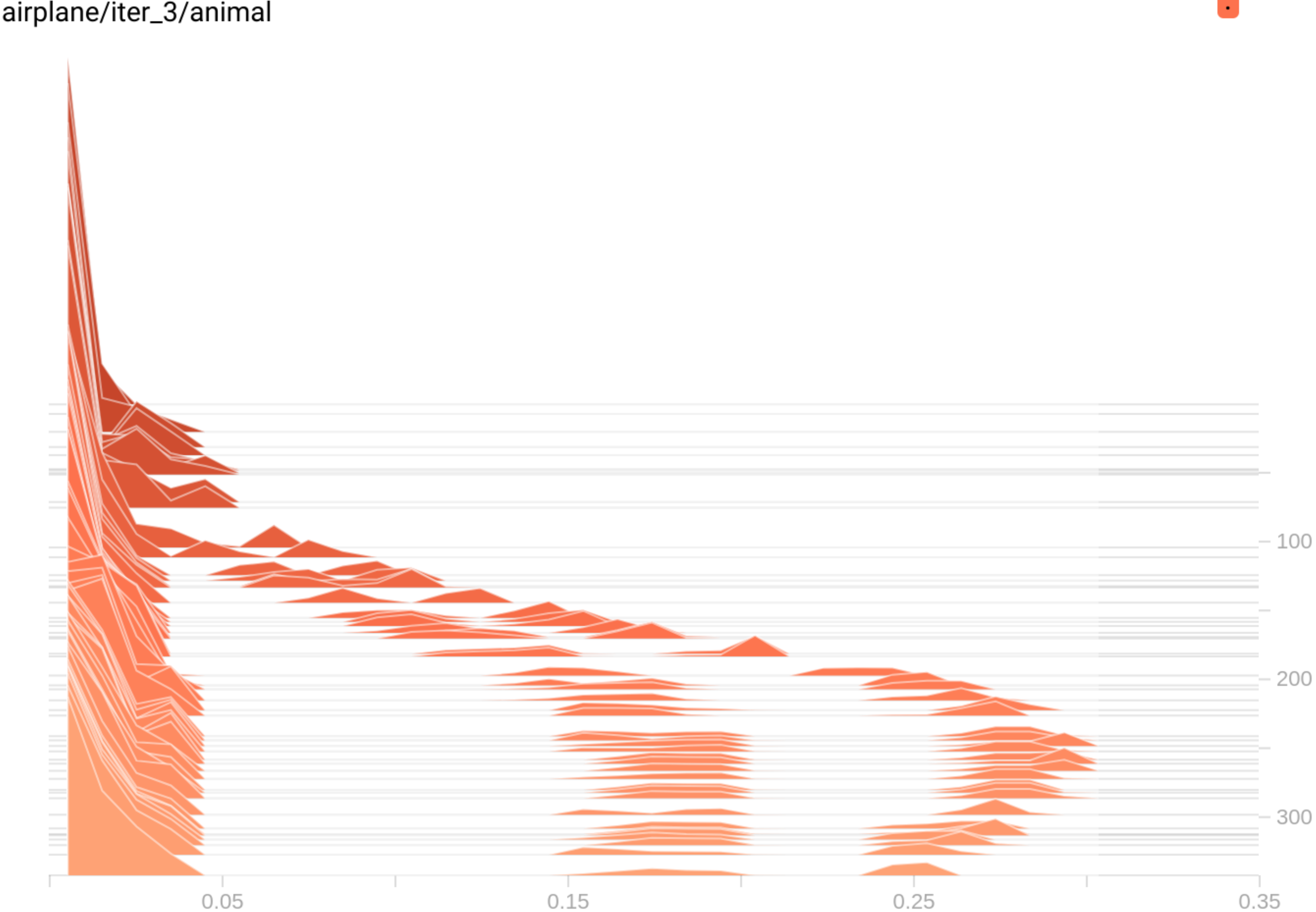}&\hspace{-1.4em}
\includegraphics[trim={0 0 0 50},clip,width=0.19\textwidth]{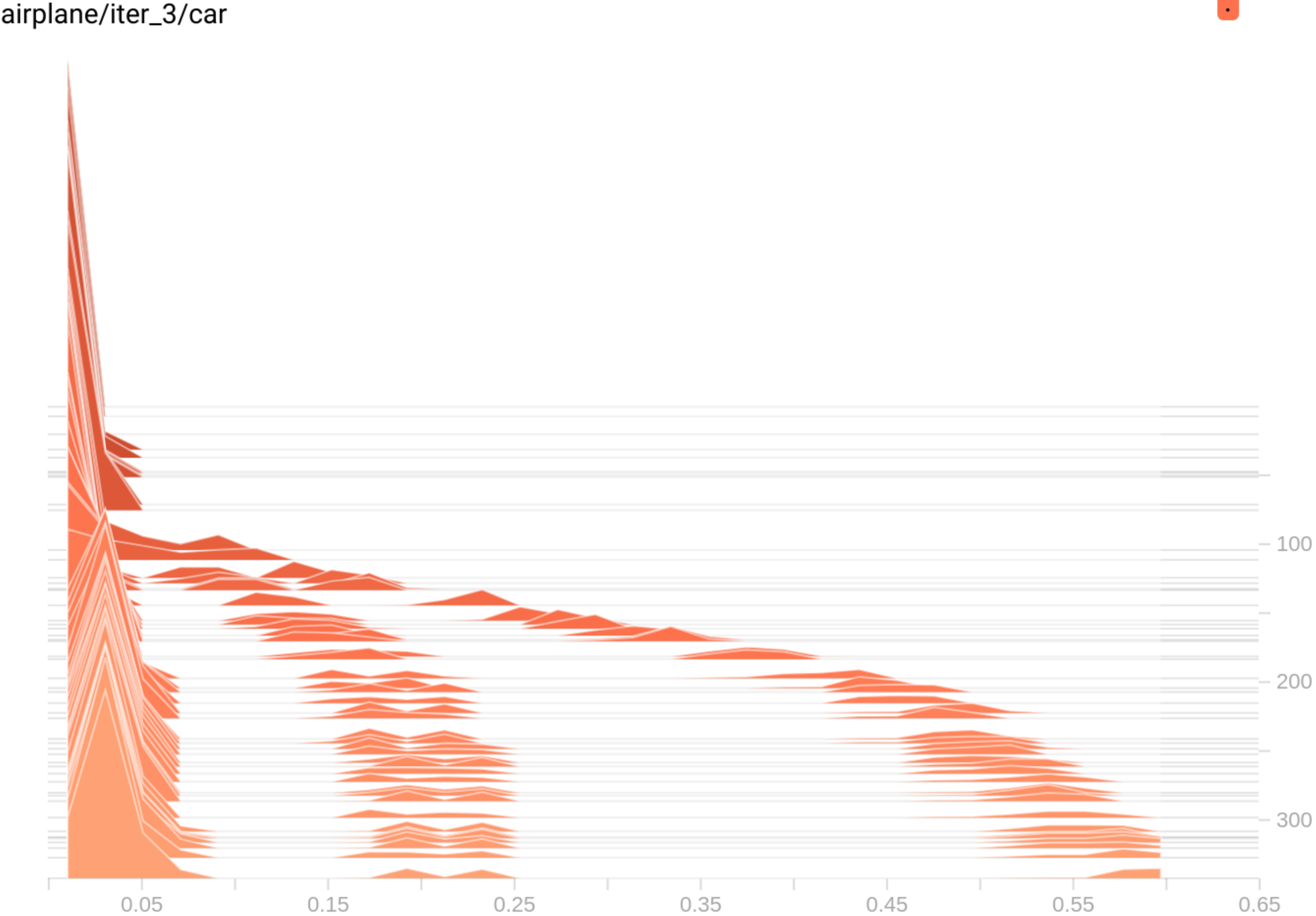}&\hspace{-1.4em}
\includegraphics[trim={0 0 0 50},clip,width=0.19\textwidth]{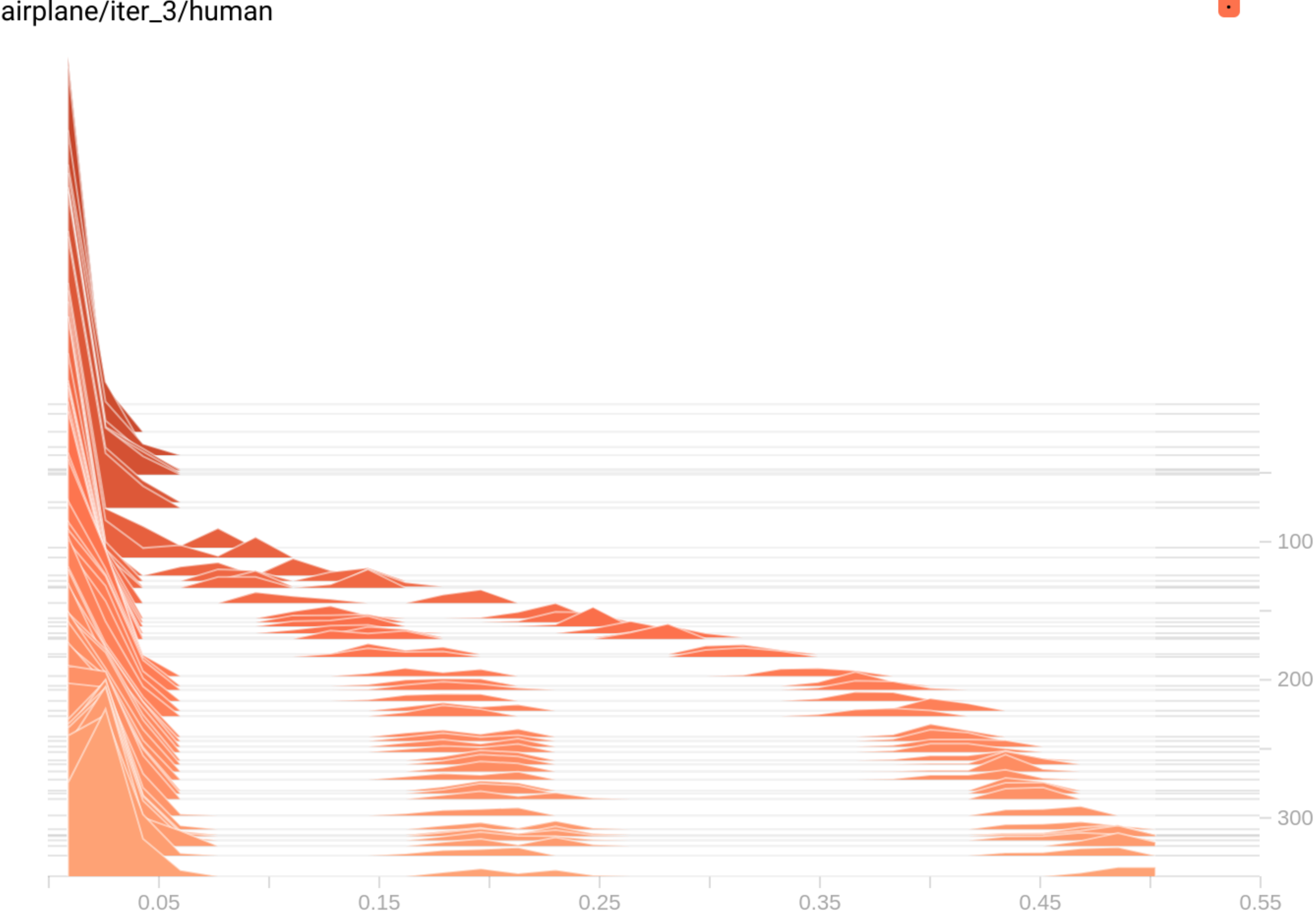}&\hspace{-1.4em}
\includegraphics[trim={0 0 0 50},clip,width=0.19\textwidth]{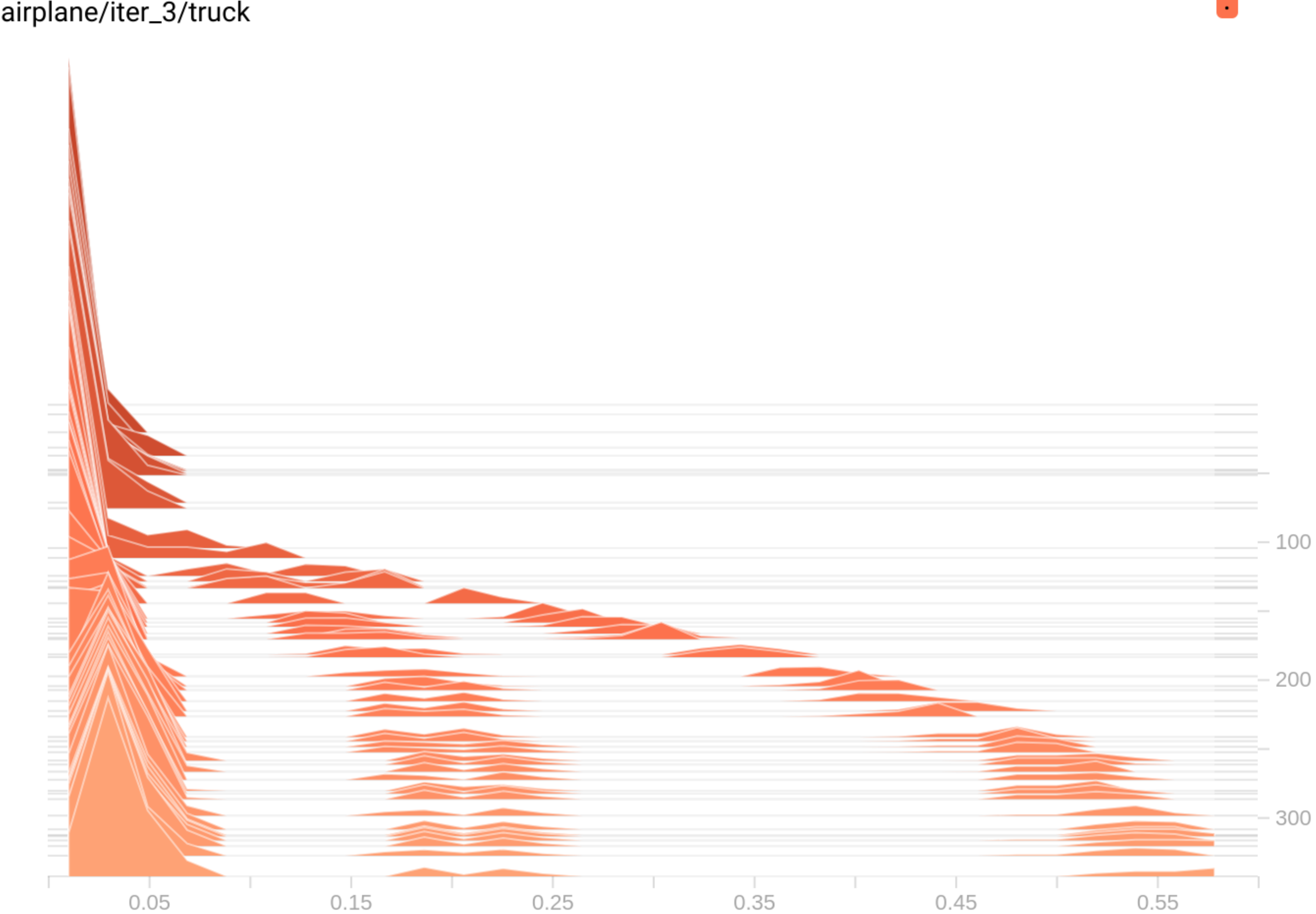}\\
\end{tabular}
\caption{Histograms of the squared distances (X axis) between votes $\mathbf{V}_{j|i}$ averaged over all \textbf{airplane} images in the smallNORB dataset, and each of the all 5 class capsules $\mathbf{M}_j$ throughout training (epochs on Y axis). Variational Bayes Routing iterations 1-3 are depicted per row, and each column represents a different class capsule. As can be seen above, the average votes from the airplane images learn to agree with the airplane class capsule during training, and therefore the discrepancies between the votes and the target capsule parameters increasingly gather around $0$ over time, more so than the other class capsules.
}
\label{airplane hist}
\end{figure*}
\begin{figure*}[t]
\settoheight{\tempdima}{\includegraphics[width=.19\linewidth]{example-image-a}}%
\centering\Large{\textbf{Car}}\par\bigskip
\centering\hspace*{-2em}\begin{tabular}{cccccc}
&\hspace{-1em} airplane &\hspace{-1.4em} animal &\hspace{-1.4em} car &\hspace{-1.4em} human &\hspace{-1.4em} truck\\
\rowname{iter 1}&\hspace{-1em}
\includegraphics[trim={0 0 0 50},clip,width=0.19\textwidth]{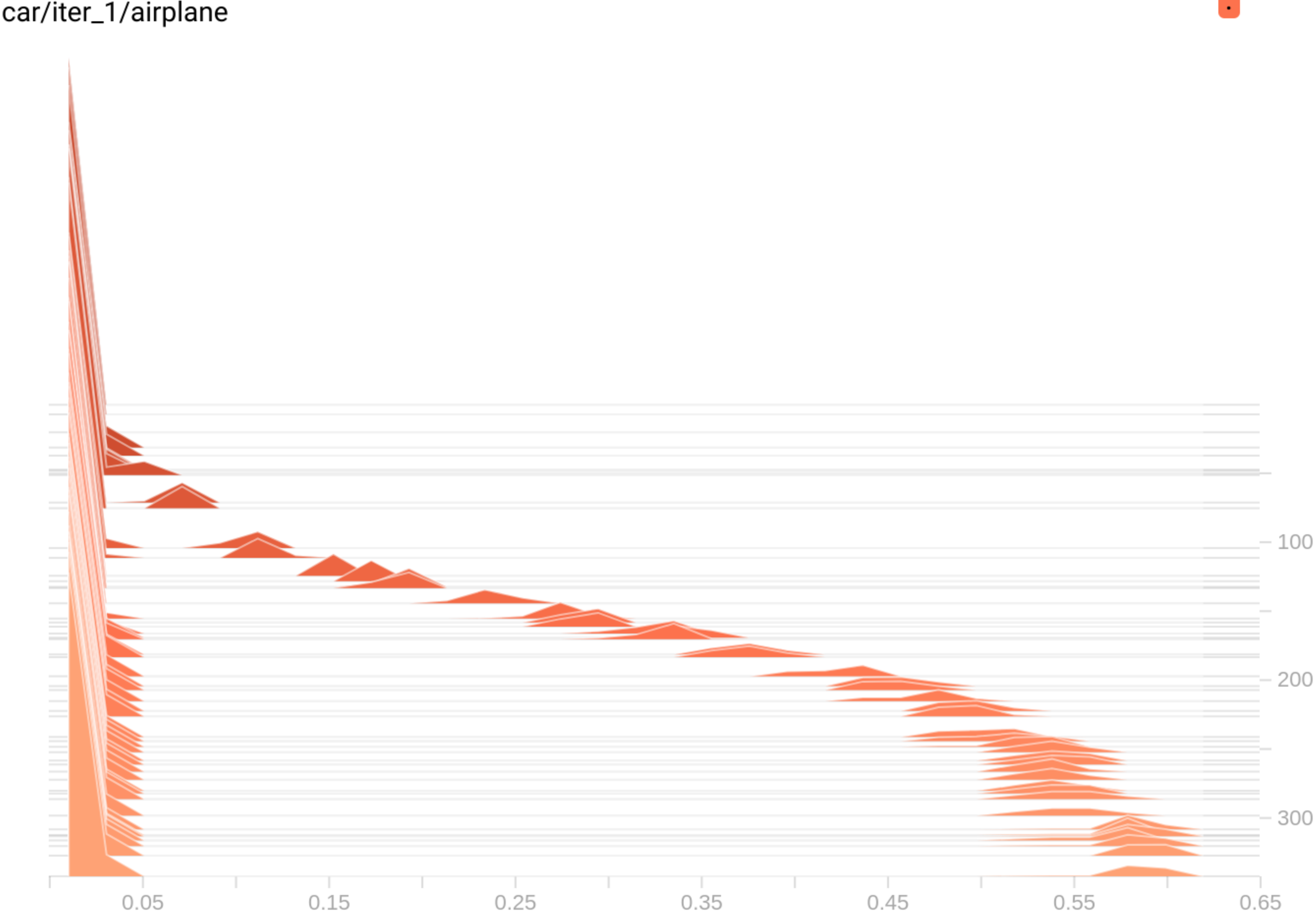}&\hspace{-1.4em}
\includegraphics[trim={0 0 0 50},clip,width=0.19\textwidth]{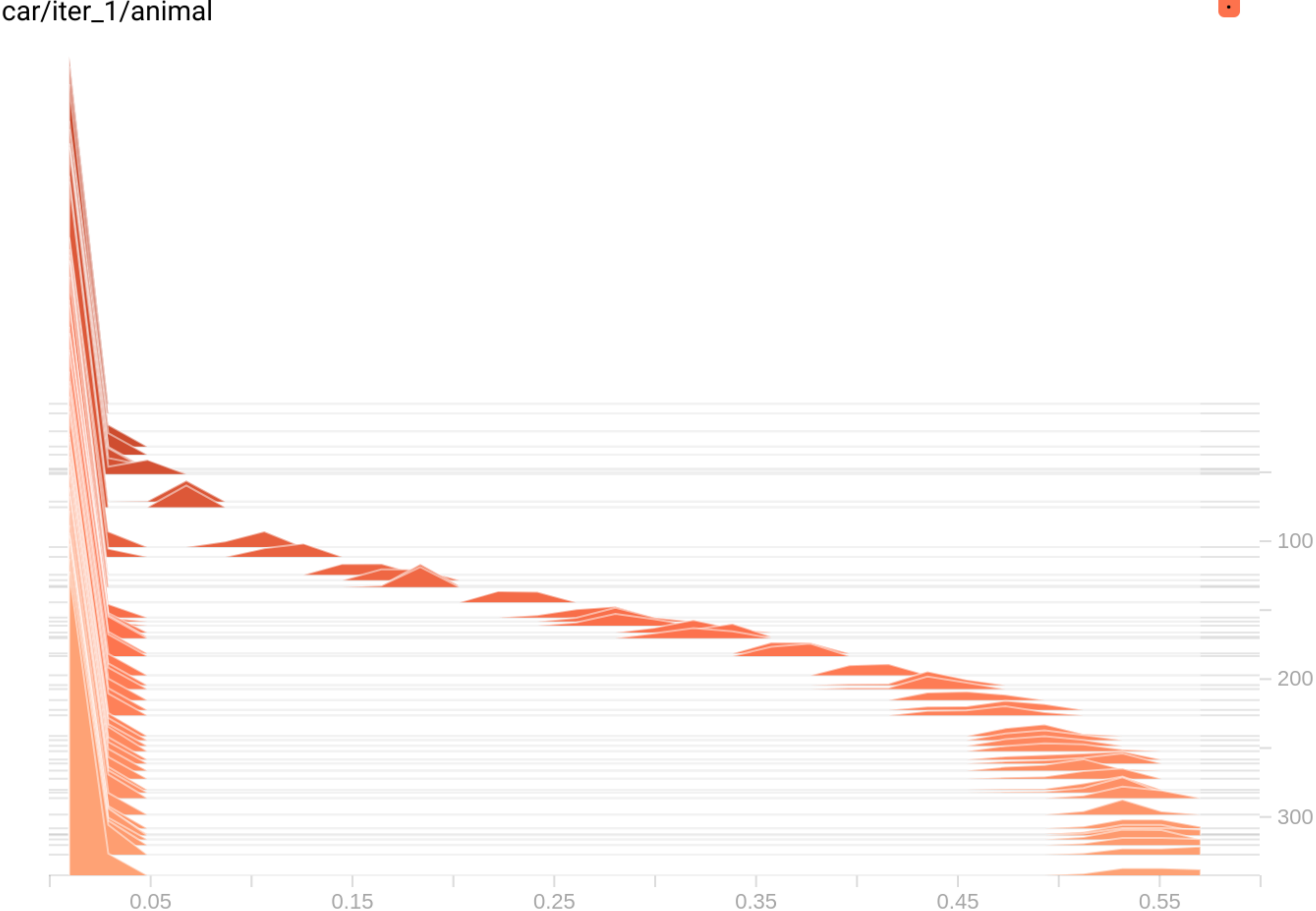}&\hspace{-1.4em}
\includegraphics[trim={0 0 0 50},clip,width=0.19\textwidth]{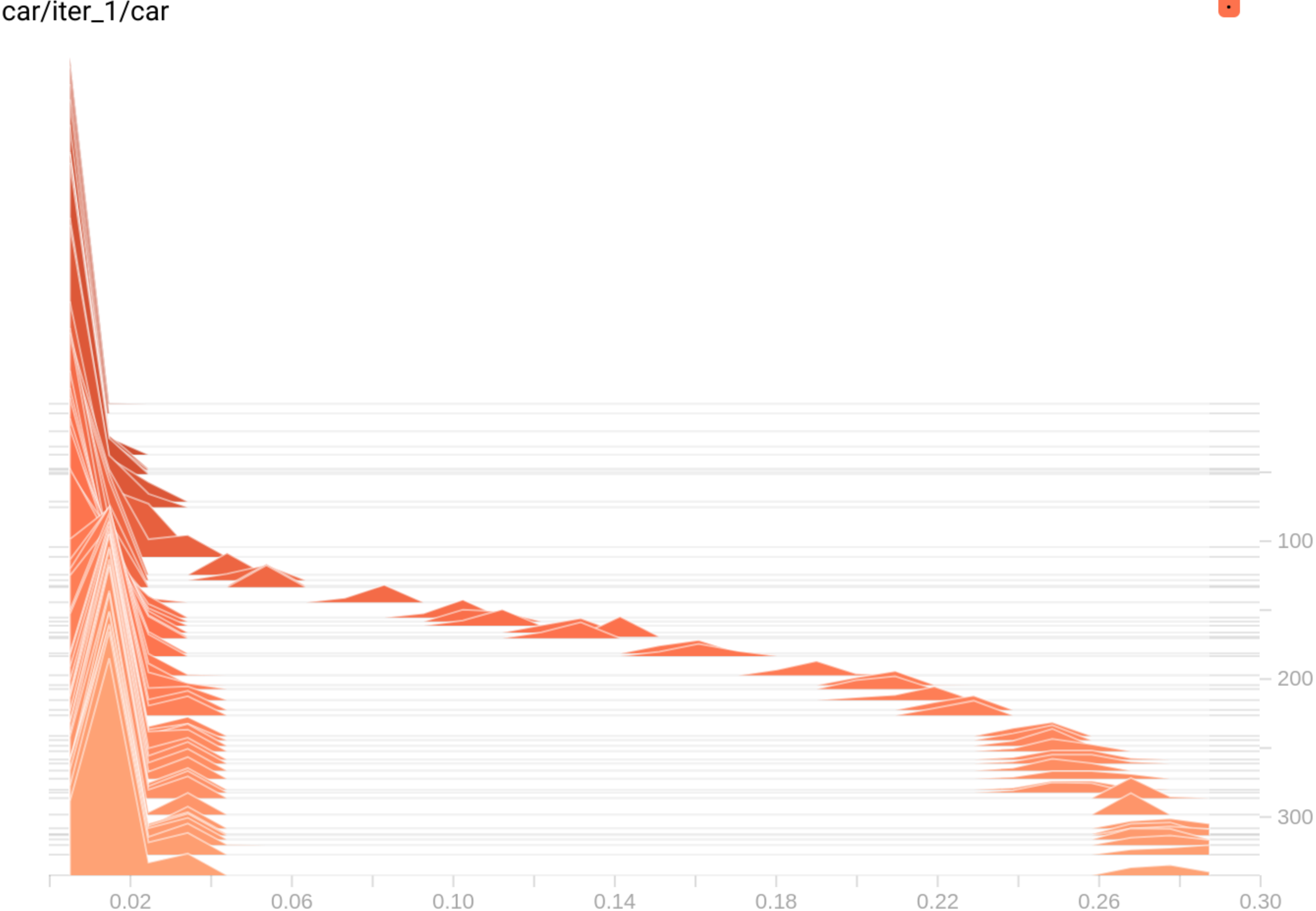}&\hspace{-1.4em}
\includegraphics[trim={0 0 0 50},clip,width=0.19\textwidth]{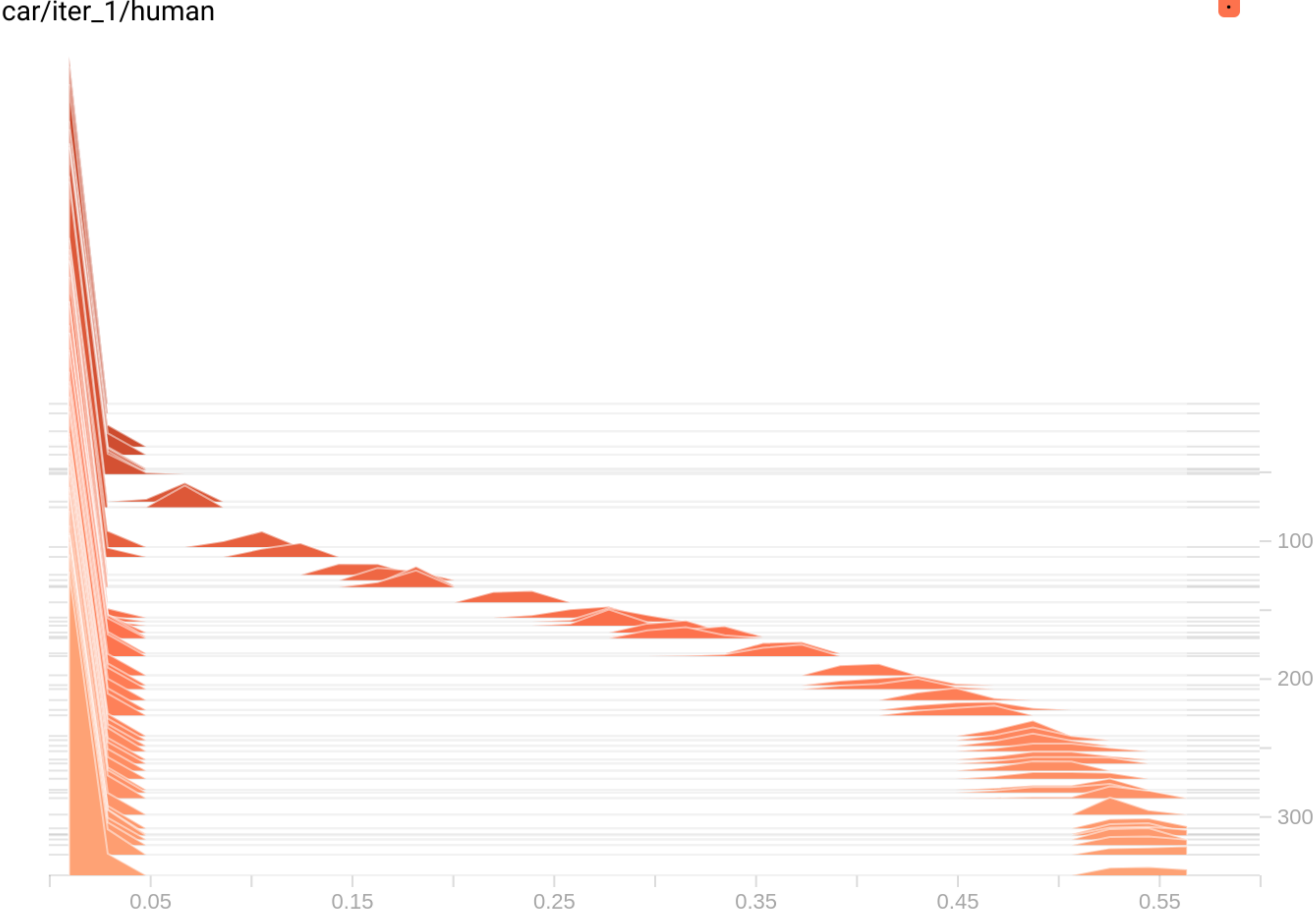}&\hspace{-1.4em}
\includegraphics[trim={0 0 0 
50},clip,width=0.19\textwidth]{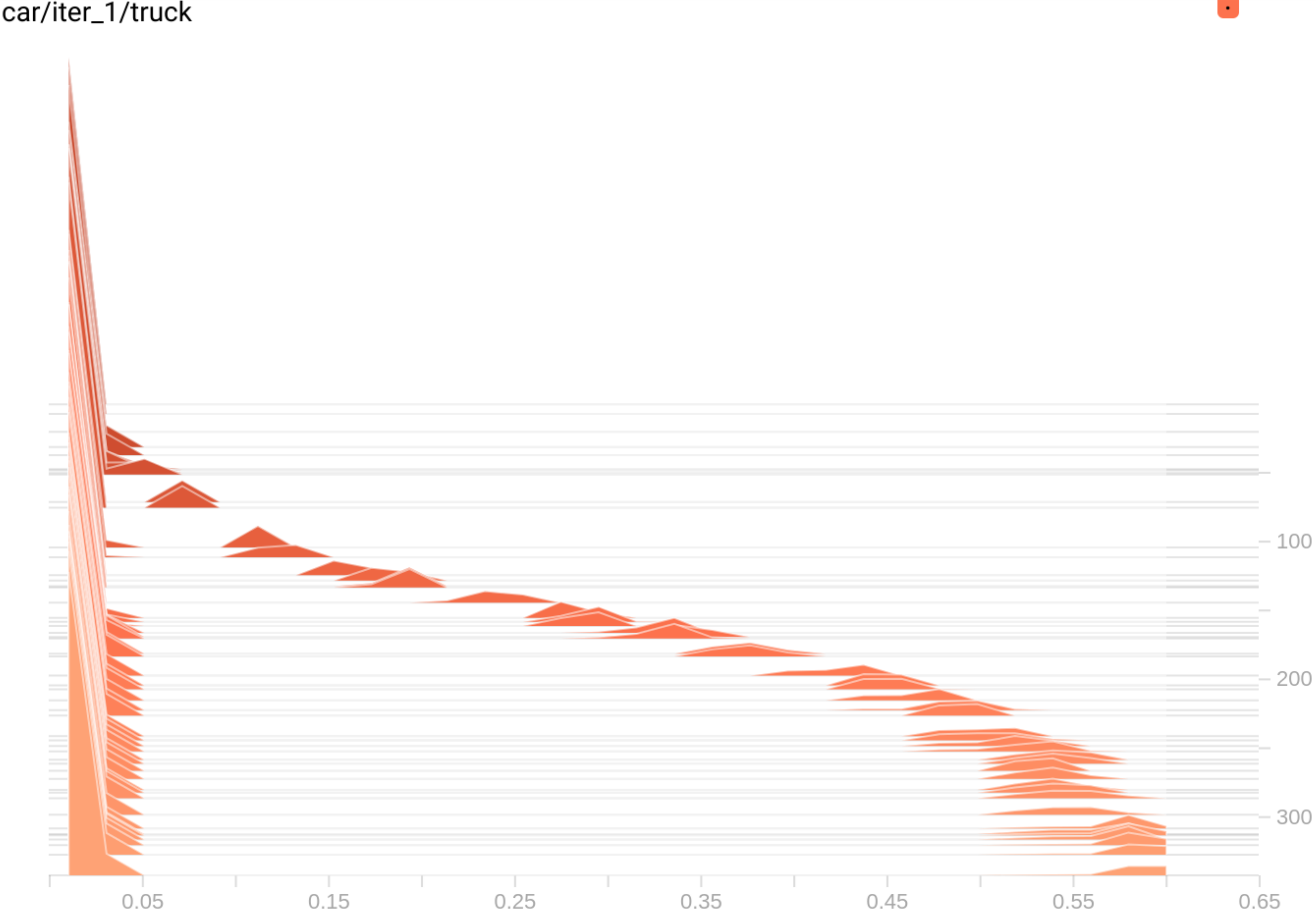} \vspace{-.4em} \\
\rowname{iter 2}&\hspace{-1em}
\includegraphics[trim={0 0 0 50},clip,width=0.19\textwidth]{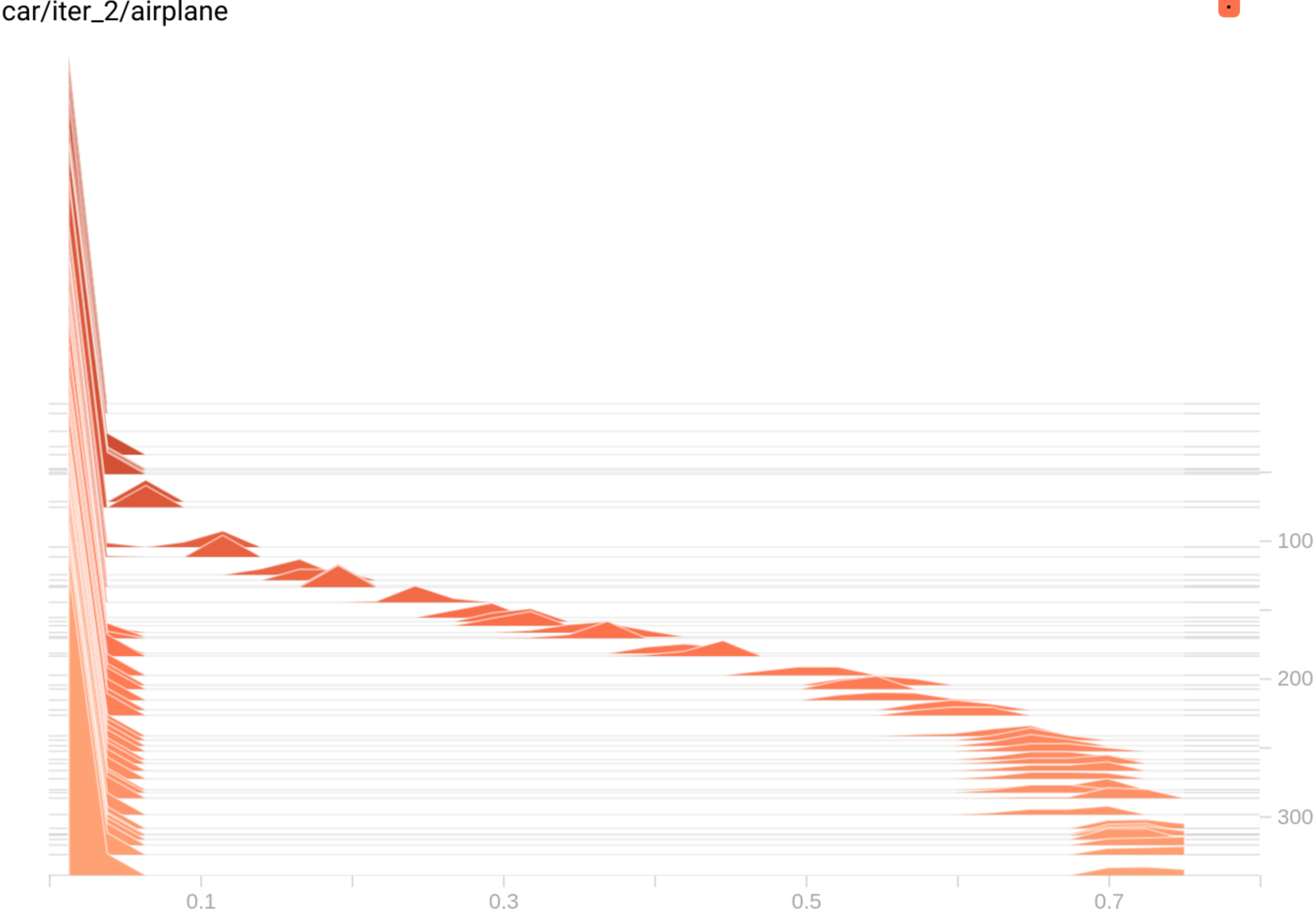}&\hspace{-1.4em}
\includegraphics[trim={0 0 0 50},clip,width=0.19\textwidth]{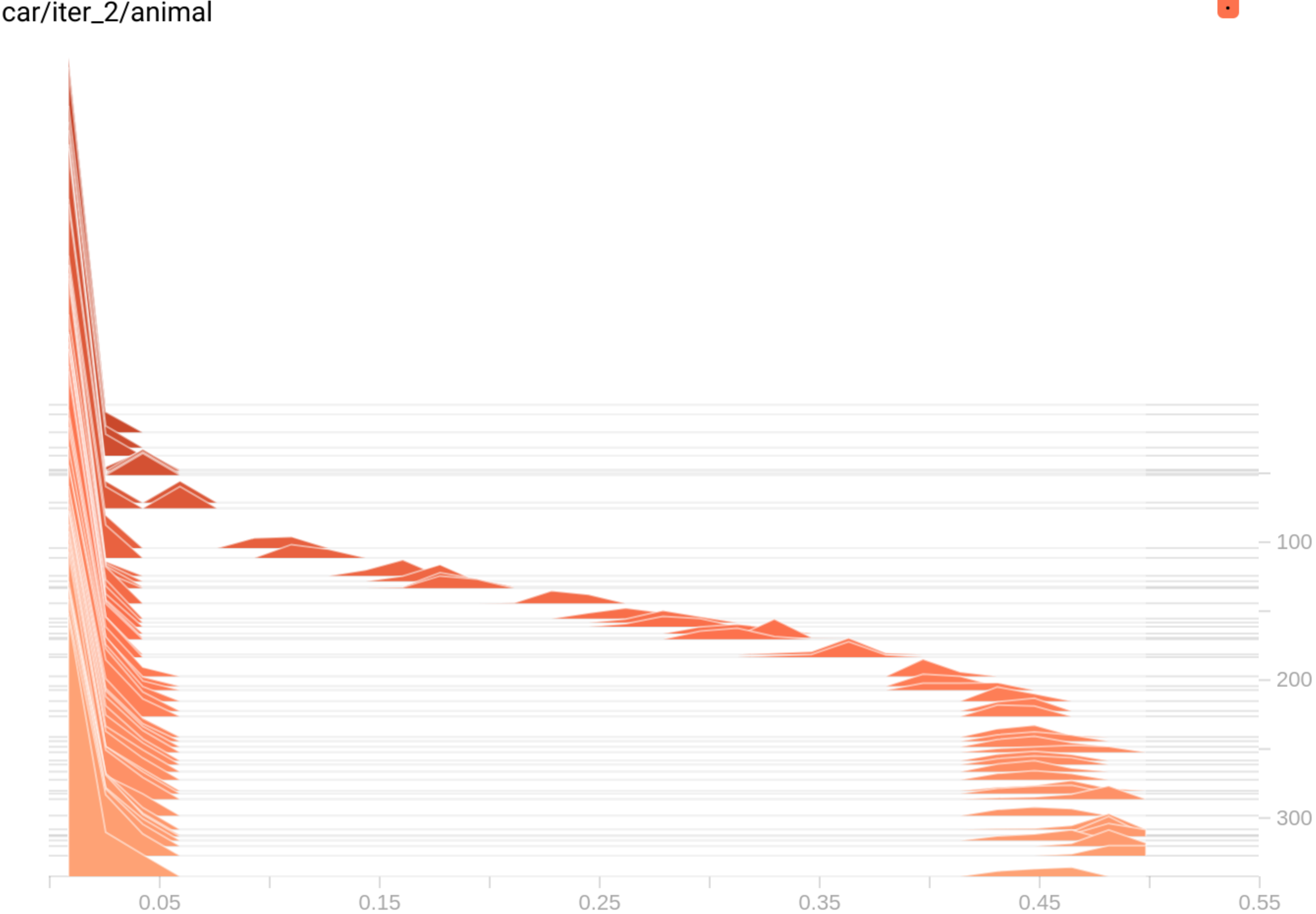}&\hspace{-1.4em}
\includegraphics[trim={0 0 0 50},clip,width=0.19\textwidth]{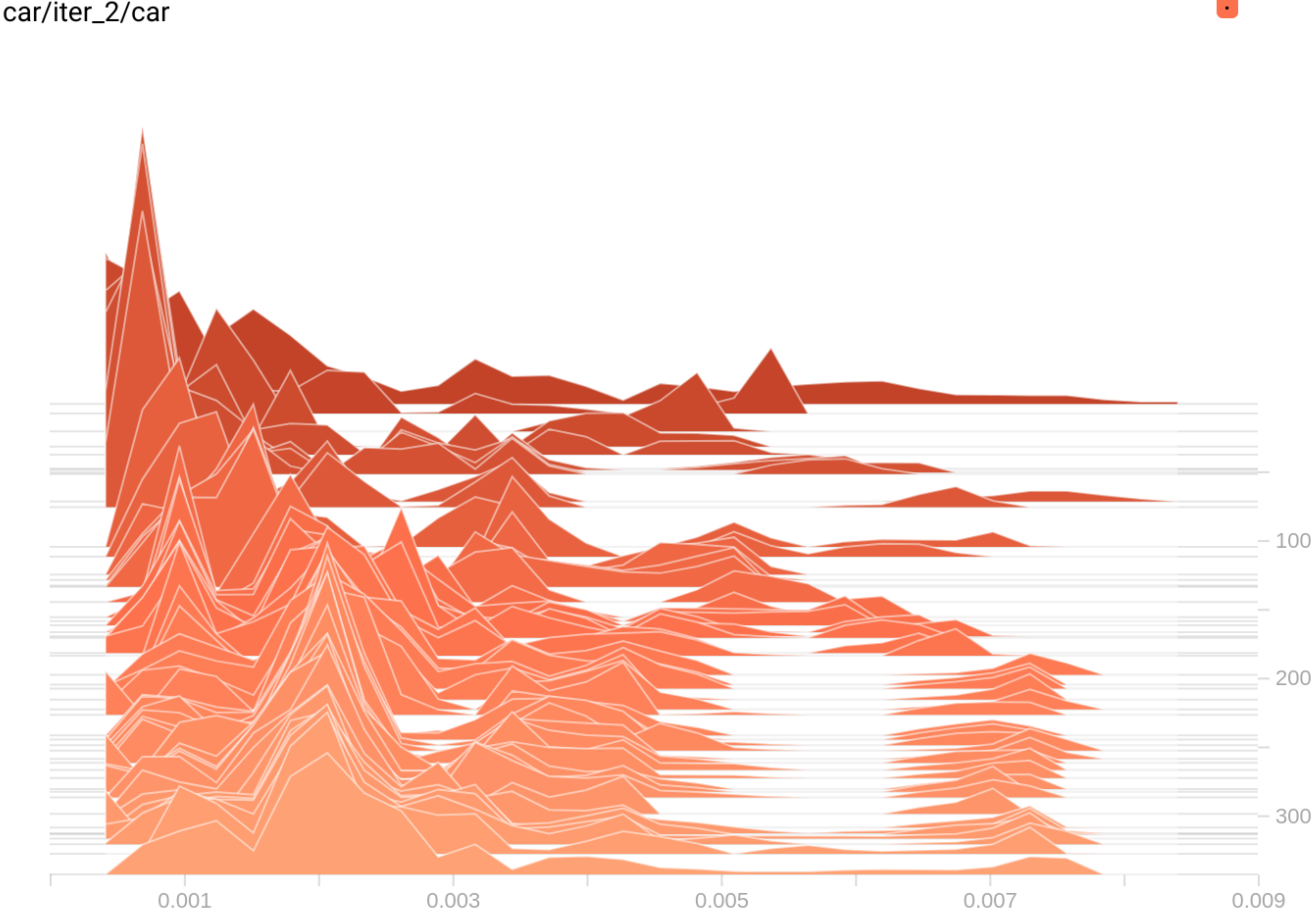}&\hspace{-1.4em}
\includegraphics[trim={0 0 0 50},clip,width=0.19\textwidth]{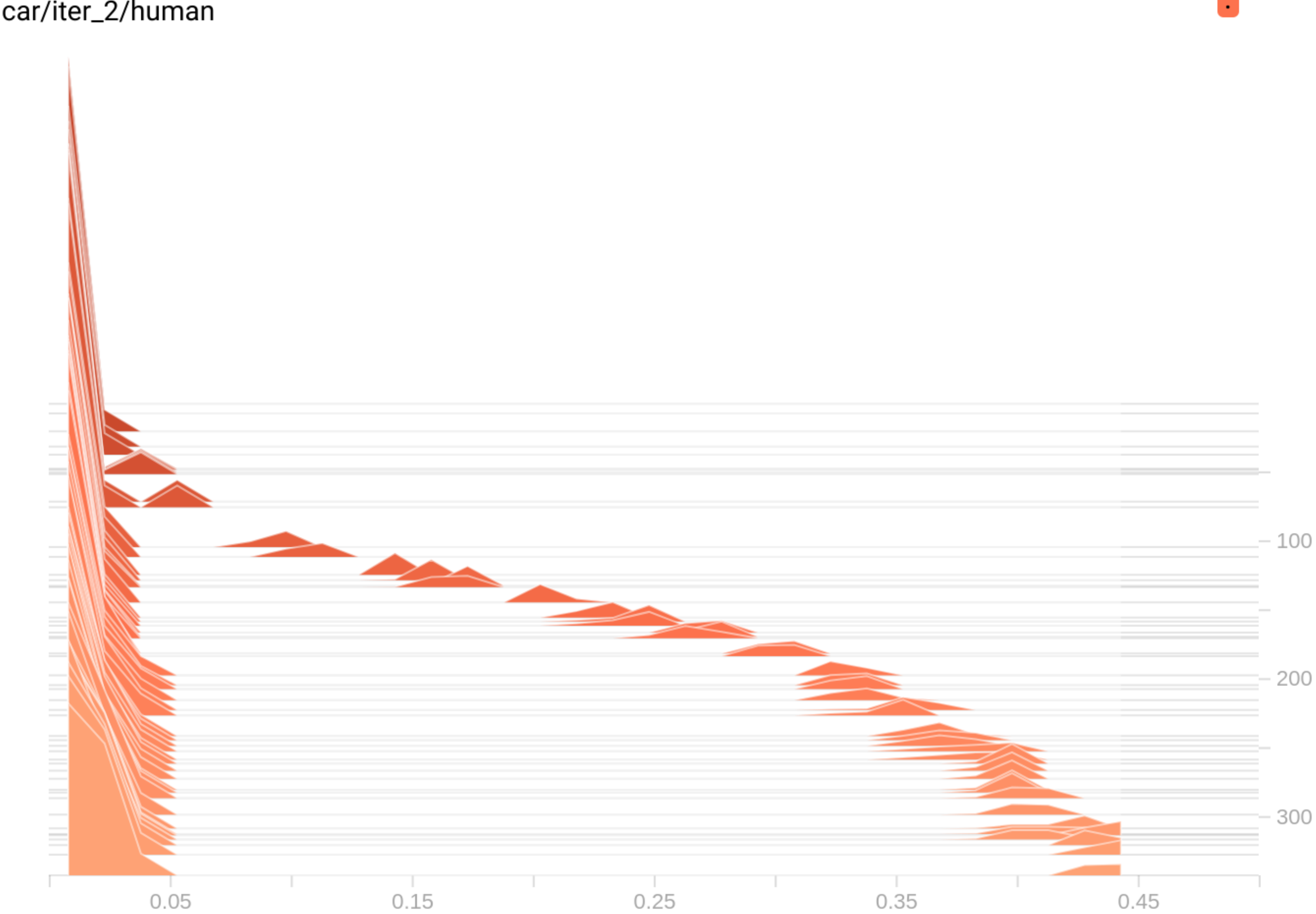}&\hspace{-1.4em}
\includegraphics[trim={0 0 0 
50},clip,width=0.19\textwidth]{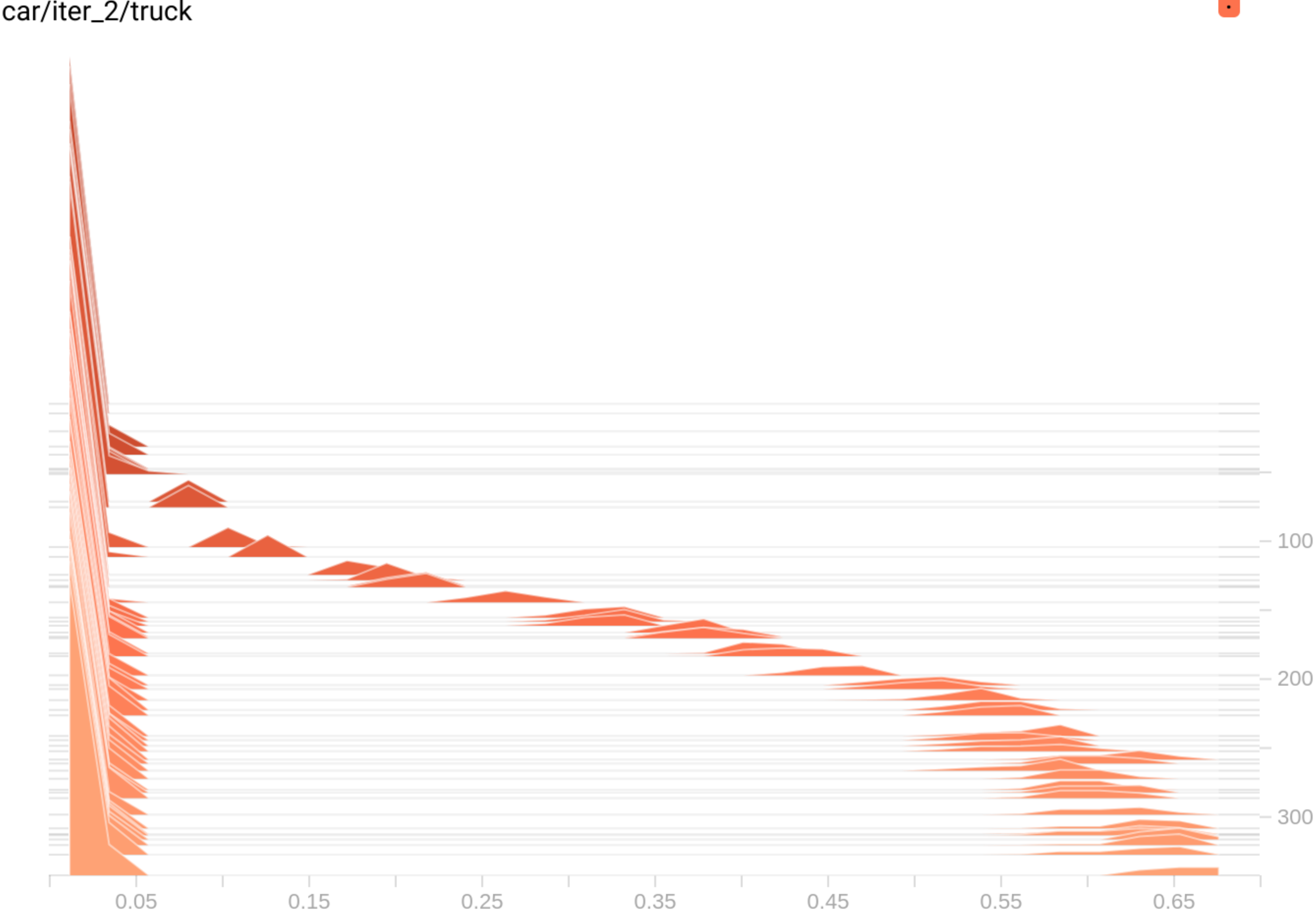} \vspace{-.4em} \\
\rowname{iter 3}&\hspace{-1em}
\includegraphics[trim={0 0 0 50},clip,width=0.19\textwidth]{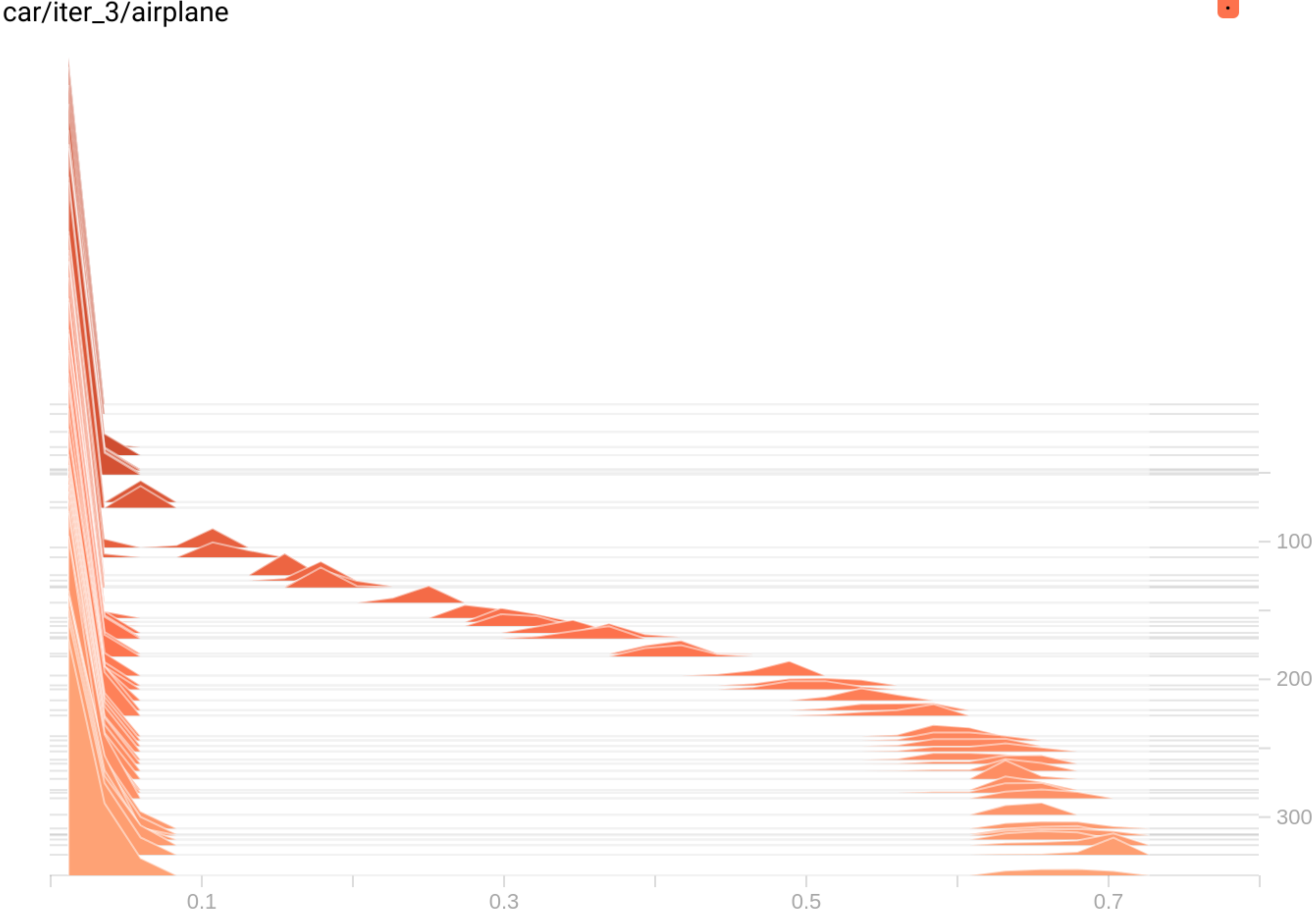}&\hspace{-1.4em}
\includegraphics[trim={0 0 0 50},clip,width=0.19\textwidth]{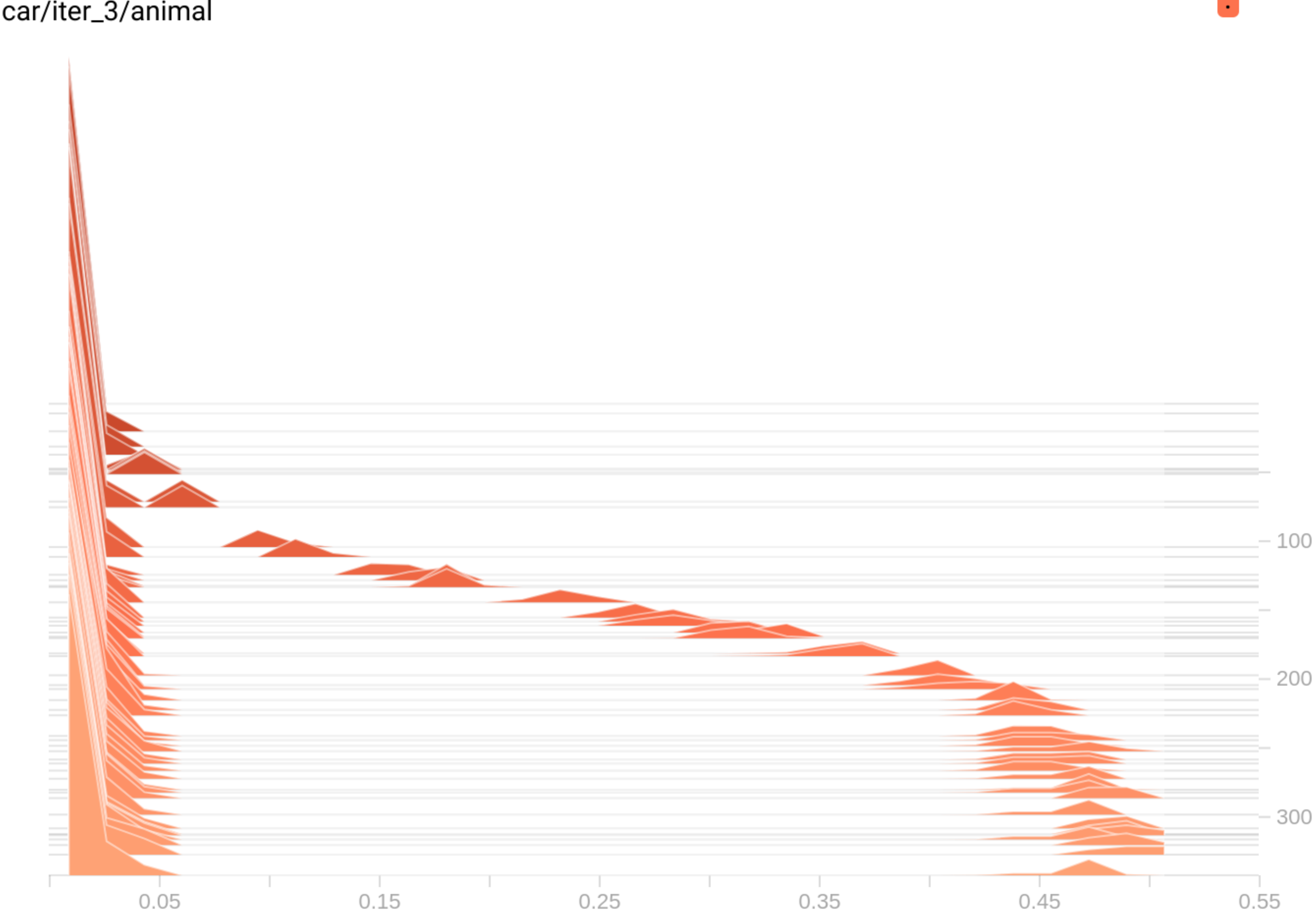}&\hspace{-1.4em}
\includegraphics[trim={0 0 0 50},clip,width=0.19\textwidth]{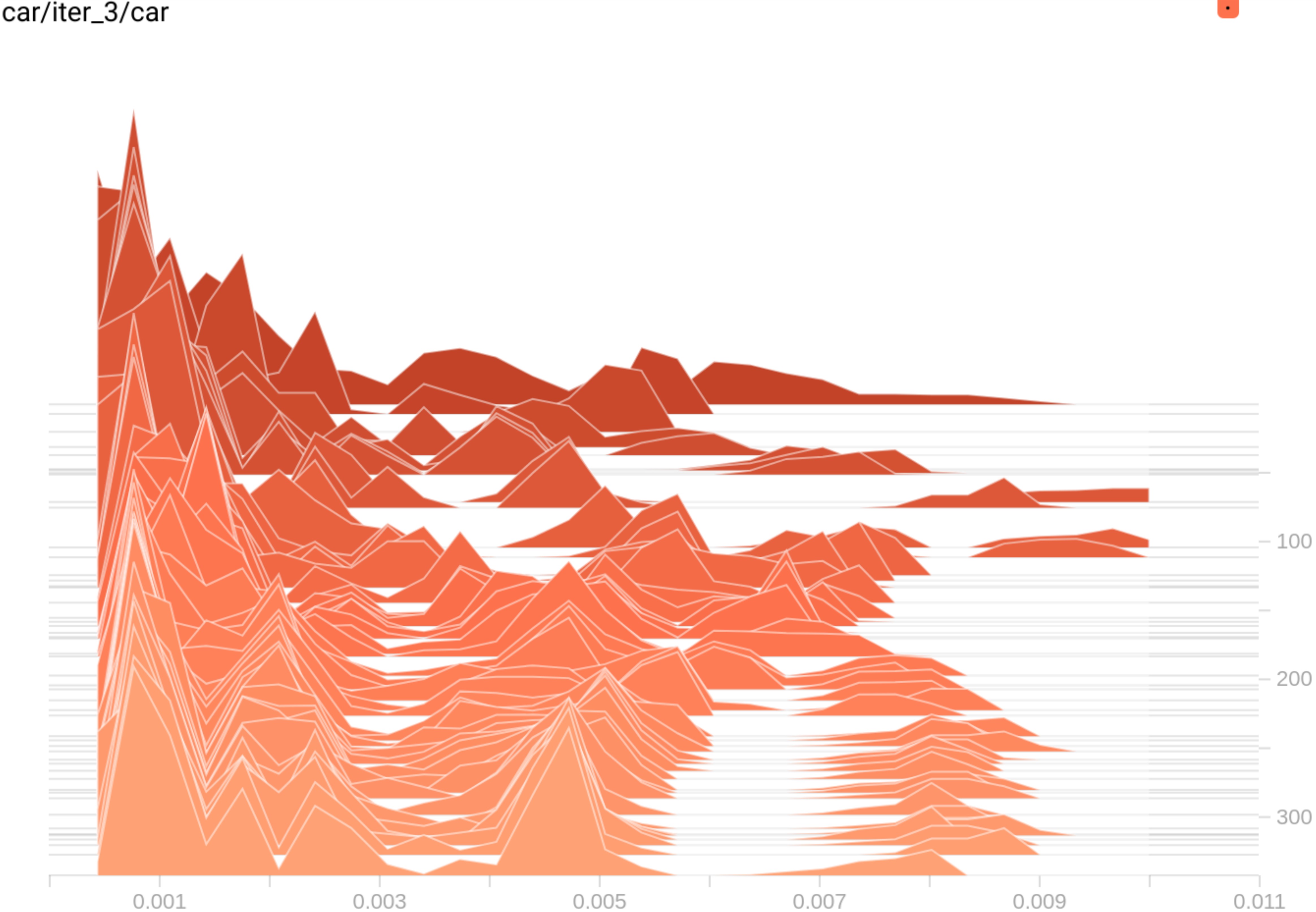}&\hspace{-1.4em}
\includegraphics[trim={0 0 0 50},clip,width=0.19\textwidth]{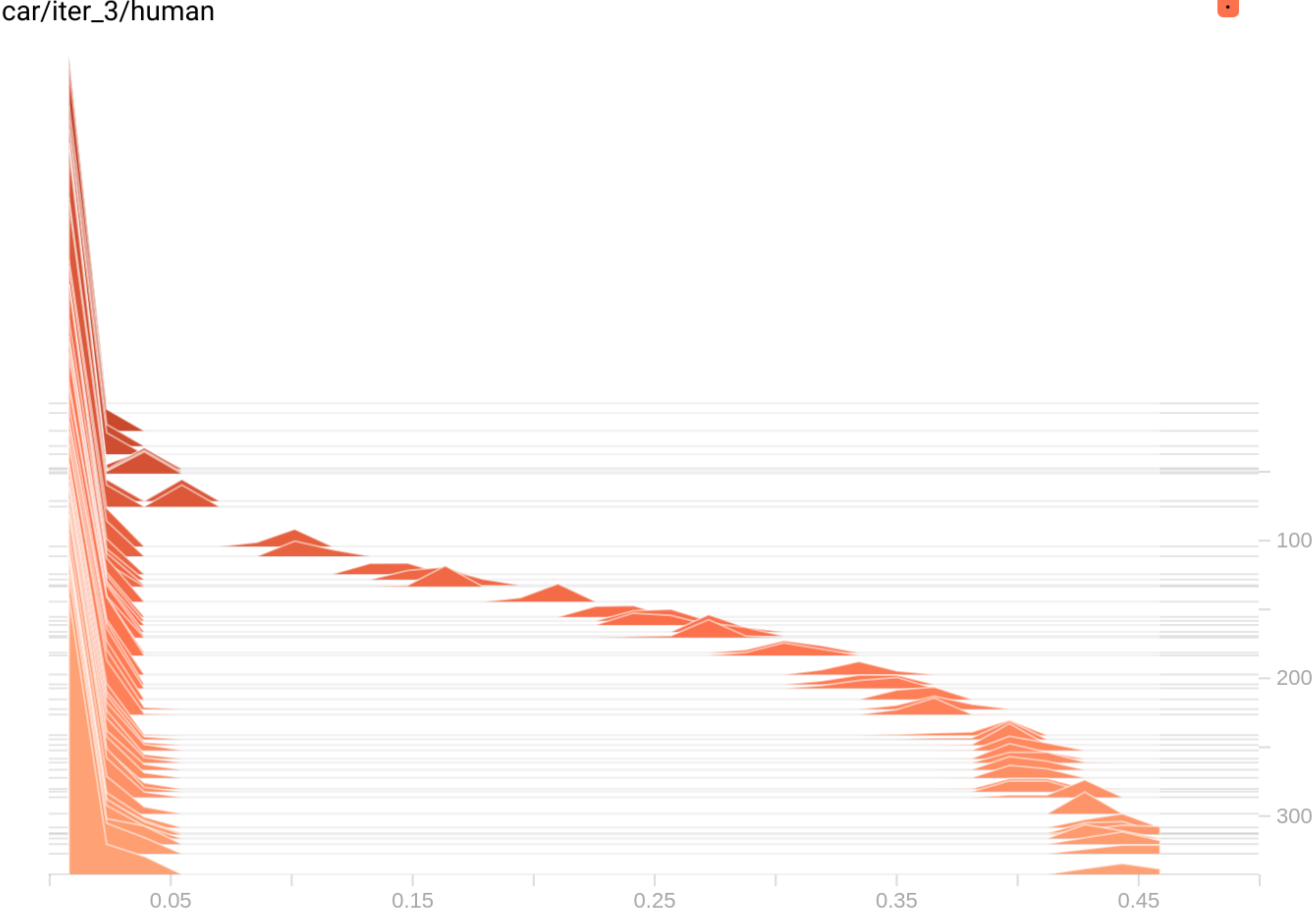}&\hspace{-1.4em}
\includegraphics[trim={0 0 0 
50},clip,width=0.19\textwidth]{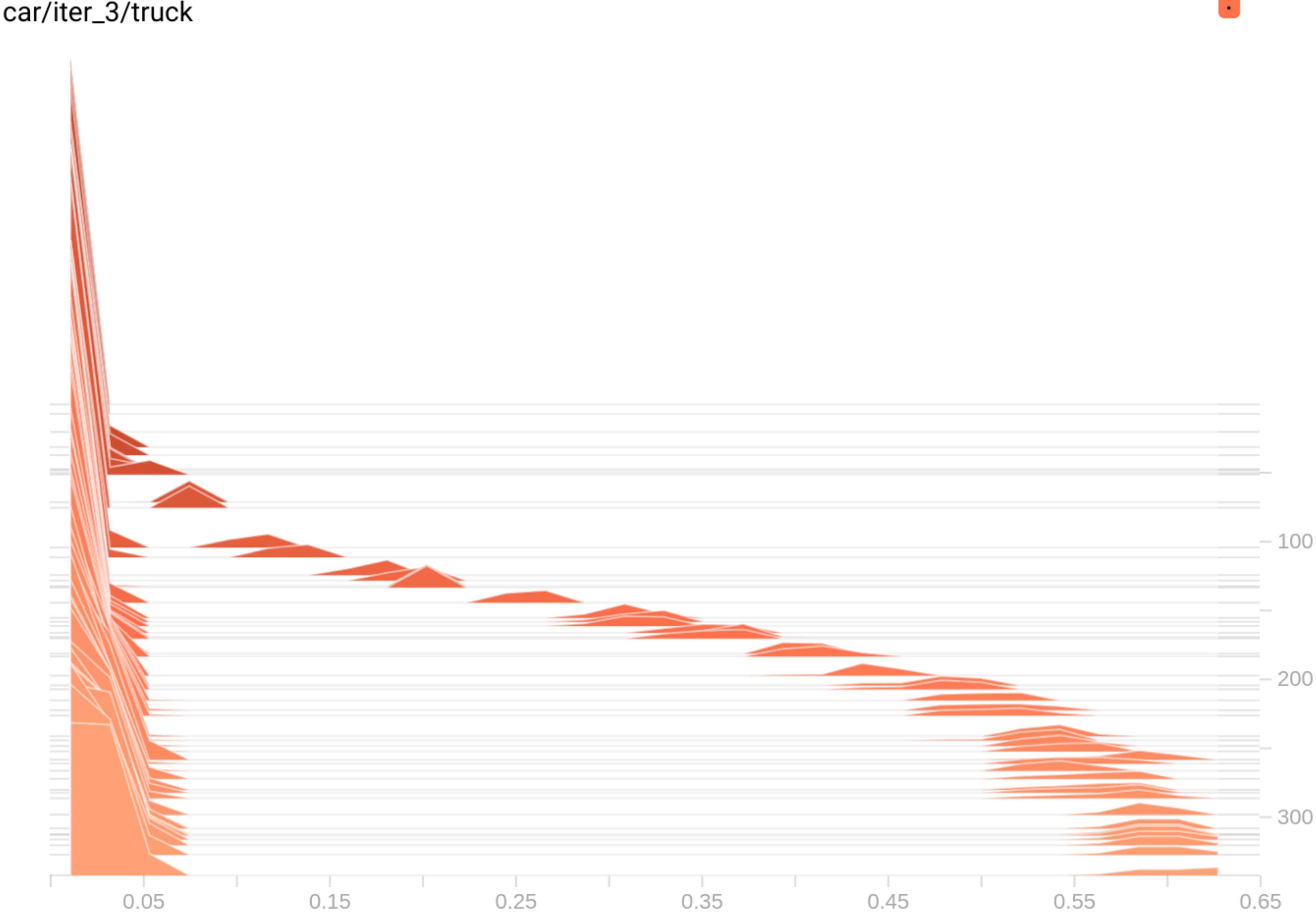}\\
\end{tabular}
\caption{Histograms of the squared distances (X axis) between votes $\mathbf{V}_{j|i}$ averaged over all \textbf{car} images in the smallNORB dataset, and each of the all 5 class capsules $\mathbf{M}_j$ throughout training (epochs on Y axis). Variational Bayes Routing iterations 1-3 are depicted per row, and each column represents a different class capsule. A very clear difference in the agreement between target (car) and non-target capsules as training progresses can be seen without inspecting the absolute distances on the X axis.}
\label{car hist}
\end{figure*}
\end{document}